\documentclass[11pt]{article}
\usepackage[utf8]{inputenc}
\usepackage[margin=1in]{geometry} 

\usepackage{blindtext}
\usepackage{xcolor}
\usepackage{subcaption}
\usepackage{booktabs}
\usepackage{multirow}
\usepackage{varwidth}
\usepackage{amssymb}
\usepackage{pifont}
\usepackage{amssymb}
\usepackage{natbib}
\usepackage{graphicx}
\usepackage[hidelinks]{hyperref}
\usepackage[noblocks]{authblk}

\usepackage{cleveref}

\usepackage{lastpage}

\begin{document}

\title{Contrasting local and global modeling with machine learning and satellite data: 
A case study estimating tree canopy height in African savannas 
}

\makeatletter
\renewcommand*{\@fnsymbol}[1]{\@arabic{\numexpr#1\relax}}

\author[1]{Esther~Rolf\hspace{.1em}}
\author[2]{Lucia~Gordon\hspace{.1em}}
\author[2]{Milind~Tambe\hspace{.1em}}
\author[2]{Andrew~Davies\hspace{.1em}}

\affil[1]{University of Colorado, Boulder}
\affil[2]{Harvard University}

\maketitle

\def  \jmlrsubmissionversion {false}
\def  \truecondition {true}

\begin{abstract}
While advances in machine learning with satellite imagery (SatML) are facilitating environmental monitoring at a global scale, developing SatML models that are accurate and useful for local regions remains critical to understanding and acting on an ever-changing planet.
As increasing attention and resources are being devoted to training SatML models with global data, it is important to understand when improvements in global models will make it easier to train or fine-tune models that are accurate in specific regions.
\ifx\jmlrsubmissionversion\truecondition{
To explore this question, we design the first study that explicitly contrasts local and global training paradigms for SatML, through a case study of tree canopy height (TCH) mapping in the Karingani Game Reserve, Mozambique. }
\else{
To explore this question, we contrast local and global training paradigms for SatML through a case study of tree canopy height (TCH) mapping in the Karingani Game Reserve, Mozambique. }
\fi
We find that recent advances in global TCH mapping do not necessarily translate to better local modeling abilities in our study region.  
Specifically, small models trained only with locally-collected data outperform published global TCH maps, and even outperform globally pretrained models that we fine-tune using local data.
Analyzing these results further, we identify specific points of conflict and synergy between local and global modeling  paradigms that can inform future research  toward aligning local and global performance objectives in geospatial machine learning.
\end{abstract}

\section{Introduction}
\label{sec:intro}

Paired with a massive amount of publicly available satellite imagery, geospatial machine learning is enabling a new paradigm of environmental monitoring. Machine learning with satellite imagery (SatML) models have been deployed at a global scale to classify patterns as diverse as land cover types \citep{Brown2022}, industrial activity at sea \citep{Paolo2024}, micro-estimates of wealth \citep{Chi2022}, and forest and tree canopy height \citep{glad,eth,meta}. 
While efforts like these significantly extend the possibilities for \emph{global-scale}  science and policy, understanding \emph{local} trends and variations in key environmental indicators remains paramount.
Many important indicators of global change in ecology and environmental sciences manifest at local or regional scales, which global-scale data do not always represent well. Moreover, data-driven policies designed to mitigate global change and its effects (e.g., policies that regulate land use, or distribute resources across populations) are
often tailored to particular localities (e.g.,  administrative regions or protected areas). 

Prior work has exposed concerns about the accuracy of geospatial machine learning models developed for global- or national-scale prediction when evaluated in local areas. \cite{Kerner2024} found that global models of land cover maps are sub-optimal at capturing regional land cover variation of agricultural land in Sub-Saharan Africa. \cite{Aiken2023} showed that satellite-based poverty prediction maps can exhibit policy-relevant regional biases across urban and rural lines when trained on entire countries. \cite{nyandwi2024local} highlighted limitations of global building and roads datasets evaluated across diverse neighborhoods in Rwanda. Similarly, \cite{gevaert2024auditing} found substantial gaps between local and global performance of building segmentation maps from SatML models, with implications for disaster response. While it is tempting to expect that advances in mapping environmental phenomena across the globe will extend to improved model performance in specific local regions, these examples suggest that this may not be the case. 

While past studies have exposed accuracy disparities in key local regions, they largely stop short of investigating the root causes and larger implications of these differences, and what they indicate about current methodologies in geospatial machine learning. Are drops in accuracy in some regions due to those regions being inherently difficult or unique places, or are they symptoms of a deeper discrepancy between the capacities of global and local models? When should we expect a model trained with an abundance of globally representative data will be accurate for a local region? And if local data are available for training, is fine-tuning a globally pretrained model a better strategy than training a local model from scratch? In order to answer these types of practical questions, we seek to develop a deeper understanding of when fundamental discrepancies can arise between local and global modeling objectives.

In this work, we devise a set of experiments to better understand if and when global mapping and local mapping use cases may constitute somewhat distinct goals for 
SatML and geospatial machine learning more generally. 
We conduct a case study of specific local ecological importance, the mapping of tree canopy height (TCH), which is also representative of a class of local modeling problem settings more generally.
Through this case study, we assess the degree to which existing SatML models and maps designed for global performance can be used to map our local area, compared with a strategy that trains a model ``from scratch'' based only on local training data.  
Our findings provide concrete evidence of a divide between local and global modeling performance in practice, and facilitate a nuanced understanding of how data and model design decisions affect this gap. 
The results of our study have implications for current best practices and promising future directions to improve possible synergies between local and global monitoring efforts in geospatial machine learning across prediction domains. 

Specifically, we ground our inquiry with a case study in developing locally-focused models of TCH using publicly available satellite imagery and spatially restricted UAV-LiDAR data from Karingani Game Reserve, Mozambique (\Cref{fig:problem_setting}). 
Recent global prediction maps of TCH built on globally available satellite data from the Global Ecosystem Dynamics Investigation (GEDI) mission  have achieved remarkable performance in mapping variation in tree canopy height across the globe, enabling the monitoring of large-scale change in vegetation extent and structure \citep{glad,eth,meta,pauls2024estimating}. 
At the same time, local variation in TCH is pivotal for studying vegetation dynamics across landscapes. Local variation in TCH has a  strong influence on spatial heterogeneity, which directly affects biodiversity through the creation of diverse ecological niches \citep{davies2014advances,coverdale2023unravelling}, predator-prey interactions \citep{loarie2013lion,davies2016effects} and spatial patterns of aboveground carbon storage \citep{aponte2020structural,noulekoun2021structural,tetemke2021species}. 
As such, there is also a separate line of work in using SatML to expand the spatial range of high-resolution TCH data that has been collected in local regions using Airborne Laser Scanning (ALS) techniques \citep{Wilkes2015,Astola2021}. 

Despite the increased use of SatML to map tree canopies at both global and local scales, there is a lack of scholarship connecting and contrasting the efforts of local modeling and global modeling for TCH mapping. This is indicative of a larger gap in research aimed at understanding how to balance (or distinguish between) the goals of local and global usability in geospatial machine learning more generally. Developing such an understanding is crucial to making efficient and sustainable progress in geospatial machine learning for environmental monitoring -- from developing global models that can be efficiently fine-tuned for local use cases, to understanding when expensive local data-collection efforts are most needed, and how best to leverage local data among other possible data sources in SatML models.

To better understand the common and distinct facets of local and global mapping paradigms, our first research question (RQ) is: \textbf{(RQ1) How necessary are locally collected reference data for local prediction, in light of existing global models and maps?}  The results of our case study indicate that local data are critical for generating ecologically relevant, accurate models in the Karingani Game Reserve region. Our locally-trained models reduce mean absolute error of existing global TCH maps by 40-69\%. Similar performance increases extend to measures that are stratified by ecologically-relevant gradients – including true canopy height, distance to rivers, and geology.  Furthermore, using globally-pretrained models  for local fine-tuning did not substantially improve model performance compared to a random initialization. Overall, the best performing model of those we compare for our task is a model trained \emph{only} with local data.

While RQ1 addresses the importance of obtaining local data to build SatML models that perform well locally, the question of how to build accurate local models remains. Developing a locally-focused predictive model involves many design decisions, including: which model architecture to use, which imagery sources and spectral layers are used as model input, and possibly even how much local data should be collected, and where. 
In order to understand the overall impact and possible co-dependencies of these different types of decisions, our second research question is: \textbf{(RQ2) Which designable factors most influence reliability of local TCH models?} We design experiments to compare the relative importance of: (i) the quantity and composition of training data sampled across ecological gradients \emph{(geographic representation)}, (ii) spectral layers of the satellite imagery \emph{(feature representation)}, and (iii) different machine learning architectures \emph{(model representation)}, on the performance of our local models. We find that these factors all have a large influence on model performance, with each factor having roughly the same magnitude of importance, and nuanced interactions between factors. 

Our case study of TCH mapping in the Karingani Game Reserve is emblematic of local mapping settings that arise across many environmental monitoring domains. Thus, our experiments have implications across domains in SatML where predictive models can be trained and deployed at many different spatial scales. 
In particular, our third research question asks: \textbf{(RQ3) What are potential points of conflict or synergies in local and global modeling efforts across application domains in machine learning with satellite imagery?} 
Examining the performance of recently proposed global TCH models, we find that recent advances in global performance do not necessarily confer performance improvements in our study domain (and in fact, some methods that improve on global performance substantially degrade local performance).
While we are careful not to draw absolute conclusions from our case study from a single region, we identify specific design decisions made in past global mapping efforts that may have hindered local usability in favor of global performance, and highlight other decisions that seem to explicitly align the efforts of global and local performance.
Importantly, many of the decisions we identify as beneficial are not specific to the TCH mapping task. 
Connecting our findings to recent scholarship in other SatML tasks, we find that several indicators of tensions identified in our case study have been evidenced in other SatML domains  (including crop classification, population density mapping, poverty mapping, and building segmentation). 
Thus, we expect that many of the trends we examine here will shed light on how we might synergize local and global mapping efforts across different prediction tasks for global monitoring with SatML. 
\section{Related Work}
\subsection{Local vs. global prediction in general and geospatial machine learning} 
The distinction between learning representations that capture local variation and those that best capture global variation has long been established, for example concerning notions of “local” and “global” learning paradigms across general manifolds \citep{Zhou2003,Saul2003,Huang2008}. There is a corresponding discrepancy in output (prediction) space as well. A model trained to optimize average performance across an entire population or distribution can fail to capture meaningful variation within sub-regions or sub-populations due to limited model capacity or finite training data \citep{rolf2022striving}, or because the model can mainly predict relative values across sub-groups of a population, but cannot reliably distinguish variation within sub-groups \citep{Aiken2023}.

Past studies have evidenced that the local-global distinction may be especially relevant for \emph{geospatial machine learning}, where the spatial nature of predictions across the globe makes the distinction between local and global prediction particularly clear. \cite{Kerner2024} evaluate existing global and regional land-cover maps in Sub-Saharan Africa, finding that global maps perform significantly worse than maps that were developed for regional use. This phenomenon extends to national-level maps as well; \cite{Yeh2020} produced national models to predict economic well-being from satellite imagery, but found that they could resolve national variation much better than variation within urban or rural areas. This result was replicated across many countries by \cite{Aiken2023}. 
Discrepancies between local and global accuracy of SatML predictions have also been highlighted in the environmental monitoring domains of population estimation \citep{kuffer2022missing}, building mapping \citep{gevaert2024auditing}, road detection \citep{nyandwi2024local} and tree canopy height mapping \citep{pauls2024estimating}.
With the increasing focus on global modeling efforts and foundation models in SatML, it is important to understand not just where errors in published models are distributed across the globe, but also to more fundamentally understand what aspects of geospatial modeling might differ for a global training and evaluation objective, compared to a more localized one. 

\subsection{Mapping tree canopy height with satellite imagery and machine learning} 
Several studies use machine learning with satellite imagery inputs to extend the spatial coverage of LiDAR-based canopy height maps in different local study regions. One thread of methodology uses locally collected LiDAR data, e.g. vegetation height from airborne LiDAR scanning (ALS), as the training labels for these predictive models. For example, \cite{Wilkes2015} used random forest models and publicly available satellite imagery to extend the range of ALS maps in east Victoria, Australia, and \cite{Astola2021} used Sentinel-2 data to generate TCH maps in Finland. Our methodology for ``local mapping'' (described in \Cref{sec:methods}) roughly follows this type of approach.

\begin{figure}[htb]
    \centering
    \includegraphics[width=\textwidth]
    {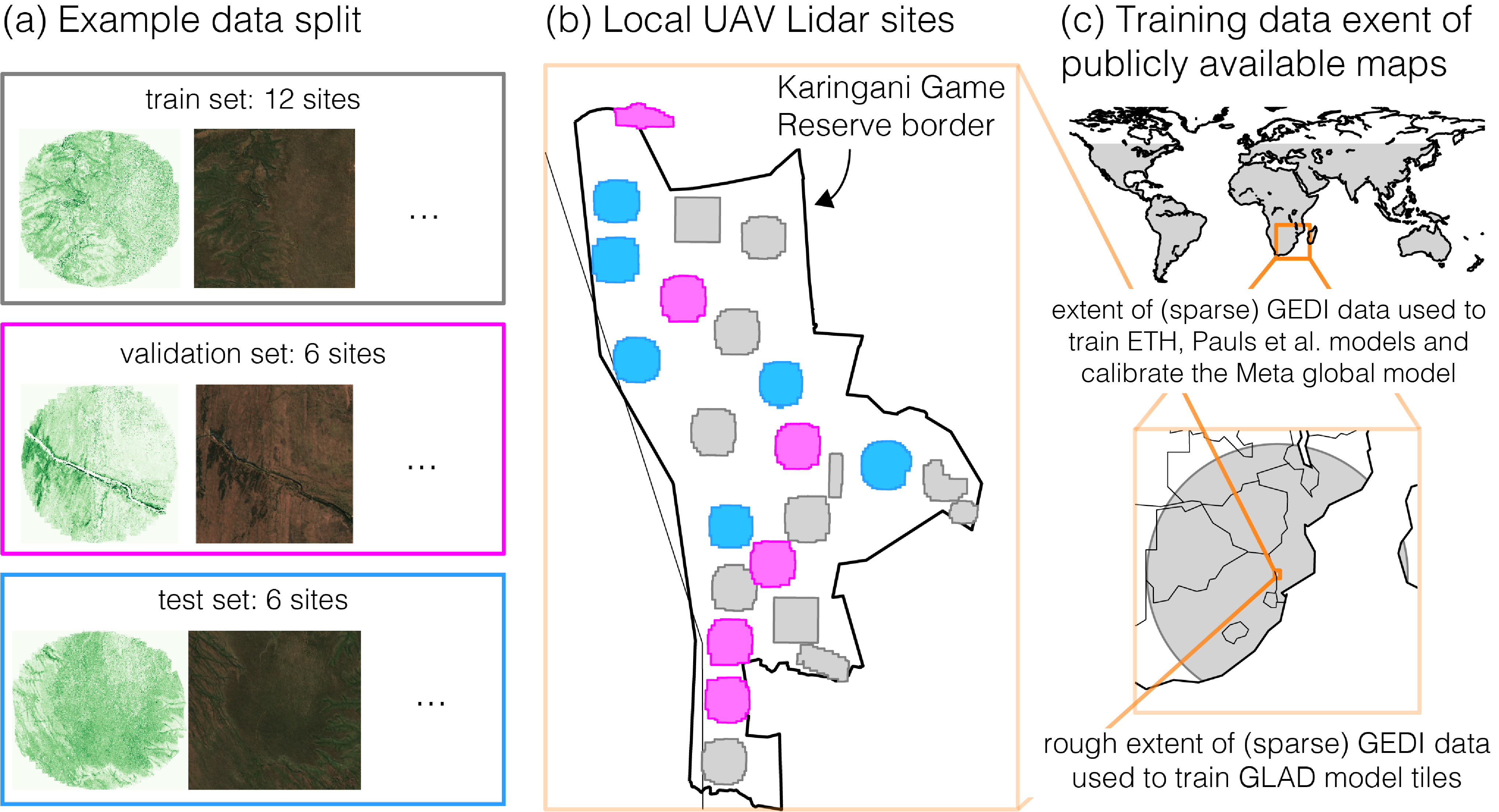}
    \caption{\textbf{Our case study simulates a common local mapping use case: we use locally-collected tree canopy height (TCH) maps derived from data collected via UAV-mounted LiDAR sensors at sites distributed across Karingani Game Reserve, Mozambique to train a predictive model, then deploy it throughout the study region}. 
    (a) The local supervised learning problem consists of high-resolution TCH labels derived from LiDAR measurements (green), paired with 12-channel Sentinel-2 imagery (3-channel visual imagery is shown here) in each site. 
    (b) There are 24 sites where UAVs were flown to collect LiDAR data, which we use to generate the local label data. Colors indicate which sites are allocated to the training (gray), validation (pink), and test (blue) sets for a given split (split 0 pictured here).  Splitting the data by site simulates only having data from the training and validation sets, using the spatially disjoint test set sites to estimate the performance that would be achieved in the other parts of the region.
    (c) Contextualizes our local study area within a global map, highlighting the rough extents of the training data used to generate the existing global TCH maps we compare to.}
    \label{fig:problem_setting}
\end{figure}

\label{sec:related_work_global_maps}
Four recent studies produce global tree canopy height maps using satellite data and machine learning. While all four maps are aimed at global coverage, each mapping effort has distinct goals that informed their design decisions:
\begin{itemize}
    \item \cite{glad} produced a 30m resolution map of canopy height across the globe for the year 2019 using a time series of Landsat satellite images. Features derived from pixelwise time series of Landsat data and digital elevation maps are used as input to bagged regression tree models, using GEDI\footnote{ While GEDI data are sparse and somewhat noisy compared to ALS or UAV-derived LiDAR data, they have the advantage of global coverage. Relevant to our study region, \cite{Li2023} assess the suitability of the GEDI RH98 data product (designed for dense forests) to African savannas around South Africa, finding that the RH98 data are reliable for canopy heights between 3-15m, with limited ability to detect shrubs below 3m.} data for label supervision. To capture regional trends, \cite{glad} train separate random forest models for each Global Land Analysis and Discovery (GLAD) Analysis Ready Data (ARD) tile (1$^\circ\times1^\circ$ region), using satellite data from the surrounding 12$^\circ$ radius (see \Cref{fig:problem_setting} for reference). Following prior convention, we refer to this as the ``GLAD'' map.
    \item \cite{eth} produced a 10m resolution global canopy height model for the year 2020 using 12-channel Sentinel-2 images concatenated with 3 channel transformations of locations (lat, lon) as input and GEDI labels as supervision. A key goal of this effort was to improve model performance for tall-height canopy regions, compared to the previous global map. \citet{eth} use a modified XceptionS2 convolutional neural network architecture. Unlike the moving window approach of \cite{glad}, \citet{eth} use a single approach globally, which consists of an ensemble of several models that is used to simultaneously output estimates of TCH and estimates of uncertainty in each region. Each model in the ensemble takes as input Sentinel-2 imagery concatenated with a 3-dimensional embedding of latitude and longitude position. Following prior convention, we refer to this as the ``ETH'' map.
    \item \cite{meta} produced a higher-resolution (1m) global canopy height model using 3-channel Maxar imagery spanning the years 2018-2020. To achieve their goal of a high spatial resolution TCH map, they pretrained a vision transformer model with dense ALS data from two regions (California, USA, and São Paulo, Brazil) then used global Maxar satellite data to extend this model across space, using  GEDI data to calibrate the pretrained model for regions across the globe. Following prior convention, we refer to this as the ``Meta" map.
    \item  Most recently, \citet{pauls2024estimating} produced a global TCH map at 10m resolution. Noting a gap in performance between regional and global TCH maps, \citet{pauls2024estimating} focused on global-scale prediction that can be competitive with local regional maps. They devised a shift-resilient loss function to account for specific structures of the GEDI measurements and included elevation data as input (in addition to Sentinel-1 and Sentinel-2 data) to improve performance in mountainous regions. We refer to this map as the \citet{pauls2024estimating} map.
\end{itemize}

The design decisions made in producing these existing global TCH maps reflect their common goal: to produce accurate \emph{global} maps of tree canopy height. 
In contrast, in this study, our aim is to (i) assess the degree to which these globally-focused maps and models can aid in the goal of producing accurate \emph{local} maps, and (ii) compare how the design decisions for global ML models and maps might differ from the design decisions for local models and maps. 

Most related to the local modeling setting of our case study, two existing studies use GEDI data to train and calibrate local TCH models. Relevant to our research questions, \cite{Healey2020} studied which spatial ranges are best for calibrating local models. Using data from across the globe, they found that local models trained on data from just a 3km radius outperformed models that were trained on data from larger radii. More recently, \cite{Tsao2023} developed local (country-level) models in Nigeria, focusing on tree height mapping for smallholder palm plantations, where performance in low-height canopy cover is important. Comparing the two globally available maps at the time (GLAD and ETH), \cite{Tsao2023} found that local calibration significantly outperformed both global maps in their study region. Similarly to \cite{Healey2020}, they find that a relatively small radius is best for locally calibrating models for predicting short trees in their region of study. It is worth noting that both of these studies use random forest models. It remains to be seen how these findings would extend to larger machine learning models that incorporate spatial information in imagery.
\section{Case Study Problem Setting: Mapping Tree Canopy Height in Karingani Game Reserve}

Karingani Game Reserve is an ecologically diverse area on the border of Mozambique and South Africa, spanning roughly 145,000 hectares. The Karingani Game Reserve was formally established in 2008 and is comprised primarily of subtropical savanna vegetation varying widely in height and density across gradients of topography, geology, soil type and mean annual rainfall. The reserve also contains a full suite of indigenous herbivore species. In addition to being an area of ecological importance, Karingani is an ideal region for a case study examining how best to map TCH in a local environment. Since the high-resolution LiDAR-based tree canopy height data we use for our case study (described in \Cref{sec:lidar_data}) has not been used to-date to train or evaluate machine learning models, our case study can simulate different possible avenues and opportunities for producing a new map of TCH in a local region.

Our case study examines how to map tree canopy height across the Karingani Game Reserve using the different resources available (a limited amount of locally collected high-resolution TCH labels (\Cref{sec:lidar_data}), publicly available satellite imagery (\Cref{sec:satellite_data}), and existing maps and models (\Cref{sec:related_work_global_maps}) that have been produced for mapping TCH at a global scale). While we are of course interested in  generating accurate maps of this region, the primary focus of our study is to understand more generally which of these current resources are most helpful for creating accurate local maps, and to identify additional resources or techniques that might aid in local mapping efforts.  

\subsection{Local LiDAR-derived tree canopy height labels}
\label{sec:lidar_data}
Our local label data used for model training and evaluation span a study area of 24 sites in the Karingani Game Reserve (\Cref{fig:problem_setting}b), collected in May and early June 2021. The 1m resolution reference tree canopy height (TCH) maps are derived by applying a canopy height model (as detailed in \cite{boucher2023flying,reiner2023more}) to LiDAR sensor data acquired by flying an unoccupied aerial vehicle (UAV) across the study sites. Of the 24 sites, 22 sites span roughly 21 km$^2$ and 2 span roughly 7 km$^2$. For context, the LiDAR data collection via UAV flight took roughly 1 day per site, including travel time. 

In contrast to the globally available (but sparse) GEDI data that past studies have used to build and extend TCH models, our local LiDAR-derived TCH labels are highly accurate ground-referenced data. Thus, we consider these data to be a valuable resource for evaluating model performance in this region. Moreover, understanding the performance gain achievable when training with local TCH data will help inform the value of acquiring local LiDAR data by flying UAVs in additional targeted regions in the future. 

\subsection{Extending the range of existing label data with machine learning and Sentinel-2 satellite data}
\label{sec:satellite_data}
While the LiDAR-derived data spans sites across the Karingani Game Reserve, there are still significant gaps in the coverage of the existing TCH data (roughly two thirds of the Karingani Game Reserve are not covered by past UAV surveying efforts). In our case study, the goal is to train machine learning models that take publicly available satellite imagery as input to extend the range of the existing TCH data. 

As input to our locally-trained machine learning models, we downloaded three Sentinel-2 satellite data tiles captured at a date near to the collection of the LiDAR data (5/13/21), which span the study area and have low cloud cover percentage over the 24 sites. Since the Sentinel-2 data has roughly 10m spatial resolution, we aligned these images with coarsened 10m-resolution LiDAR-derived TCH using cubic resampling. 
We chose to restrict our local modeling efforts to mainly 12-band Sentinel-2 satellite imagery for simplicity of comparison. Past global-mapping approaches have used different input imagery and layers: \cite{glad} use Landsat imagery, \cite{eth} append location embeddings to Sentinel-2 imagery, \cite{meta} use proprietary Maxar imagery, and \cite{pauls2024estimating} use Sentinel-1 and Sentinel-2 data and digital elevation maps. We investigate the effect of input layer choices on model performance in \Cref{sec:results_data_model_decisions}.

\subsection{Site-stratified training and evaluation splits simulate conditions of model use outside training data sites} 
To reflect our case study setting of applying the trained SatML models to fill in gaps where high-fidelity TCH data are not available, we employ a spatially stratified train/validation/test split of our data. Specifically, we split the 24 training sites into three distinct sets: training (12 sites), validation (6 sites) and testing (6 sites), as depicted in \Cref{fig:problem_setting}a,b. For each split, no test sites are seen during the training step or the model validation step in which we pick model parameters and hyperparameters. 

When we train, validate, and evaluate on spatially disjointed sets of training sites in this manner, we are essentially simulating the intended conditions of our case study \citep{rolf2023evaluation}. Specifically, we simulate training our model in regions where we have data, then deploying our model to a separate set of sites that were not used in training or validating the model. Using the spatially disjointed test set ensures that the performance we evaluate more faithfully represents our intended use case in the Karingani Game Reserve. 
We construct four such spatial data splits, where each LiDAR collection site appears in exactly one test set and exactly one evaluation set. By analyzing performance across the four splits, we can assess the variation among the models across different data settings, characterized by the available training data and target regions for model use.
\section{Methods}
\label{sec:methods}
\subsection{Methods overview}

Our experiments in \Cref{sec:results} will reflect a range of different possible methodologies for predicting tree canopy height (TCH) in a local region. Since several global TCH maps are available, perhaps the most straightforward way to produce a map of TCH in a local area is to take one of the existing global TCH maps, crop and align it to the extent of our study area, and use that subset of the global map as our predictions. This is our primary methodology to ``evaluate existing global maps'' in \Cref{sec:results_existing_maps}.

An alternative approach would be to use the limited local labels to train a new model specifically adapted to the local region. Here, we distinguish between ``local-only'' models, which we train from scratch using only labels and satellite imagery from the local region (here Karingani Game Reserve), and ``globally-pretrained, locally-adapated'' (or locally ``fine-tuned'') models, which we fine-tune using local data only, but where the model is initialized from a representation that has been pretrained on global data. We use the term ``global pretraining'' to encompass both supervised models trained to predict TCH across the globe as well as self-supervised models pretrained on global satellite imagery in a way that is agnostic to possible downstream tasks.  In \Cref{sec:results_finetuning}, we use our local data (\Cref{sec:lidar_data}) to train local-only models and fine-tune existing supervised and self-supervised models that have seen global satellite data, to compare the effectiveness of both approaches. Ultimately, contrasting the performance of using existing global TCH maps ``out of the box,'' adapting globally trained models using local data, and training models using only local data will enable us to address our first research question regarding the importance  of local data for local prediction. 

To address our second research question regarding the importance of different design choices for local prediction, we construct a set of experiments in which we systematically vary different data and modeling decisions and measure their effects on local performance. 
Specifically, in \Cref{sec:results_data_model_decisions}, we train local models using different amounts of training data to assess the importance of the abundance of local data in training or adapting models for local use. 
We train models with different input layers (Sentinel-2 bands and location encodings) as a way to estimate the relative effect that different data input decisions can have on model performance. 
To contextualize the relative effects of these data decisions in practice, we compare the range of performance under these different data conditions to the range of performance due to changing the machine learning model architecture. 

Our third research question involves identifying potential points of cohesion or distinction between the goals of local and global modeling in geospatial machine learning. Toward this end, we leverage our experimental results in this case study and pull in context from recent scientific reports across several application domains in geospatial machine learning for environmental monitoring. Contextualizing our in-depth analysis of local modeling from our case study amidst a broader set of trends in geospatial ML, in \Cref{sec:discussion-RQ3}, we pinpoint specific design decisions as well as general trends in geospatial machine learning that we believe warrant more research attention toward understanding and bridging the divide between global and local modeling.   

\subsection{Implementation details}
\ifx\jmlrsubmissionversion\truecondition
The following subsections describe the details of our experimental methodology. \emph{All code will be available in a public GitHub repository upon publication, and our 10m resolution TCH labels will be made publicly available to ensure reproducibility and promote future studies in local evaluation.}
\fi

\subsubsection{Machine learning model architectures}
We compare multiple neural network model architectures throughout our experiments: (1) an XceptionS2 convolutional neural network architecture used in the ETH global predictions \citep{eth}, which has 8 residual blocks each with two convolutional layers with 256 filters, (2) a U-Net architecture with a ResNet-18 backbone designed for pixel segmentation tasks, which takes the entire input (here a $64\times64$ crop of the satellite image) as context for the predictions, and (3) a small five-layer fully convolutional network (FCN) with 128 filters per layer and leaky ReLUs between layers. See \Cref{table:num_params} for a comparison of the number of parameters for each model.
We additionally compare to (4) non-parametric pixel-level forest models, where the input is the multiple channel input for that particular pixel. This pixelwise model does not incorporate any texture in the images, thus it constitutes a simple and informative baseline for machine learning with remotely sensed data.

\subsubsection{Model training}
Our experiments compare the strategies of training models only with local data and using local data to fine-tune models that have been initialized through pretraining with global data. When we train models only with local data (using no labels or imagery outside the study area), we randomly initialize model weights and then optimize the model using supervised learning. For the models with globally pre-trained weights available\footnote{We note that while the Meta model is published for use, fine-tuning it would require purchasing proprietary satellite imagery. Pretrained weights from \cite{pauls2024estimating} were not publicly available at the time we conducted our experiments.} (the ETH and U-Net models), we use these weights (which were derived using global data), and then fine-tune the models using local data. 
For the XceptionS2 globally pretrained weights, we use the published weights corresponding to the first model (model 0) in the ensemble from \cite{eth}, available at the GitHub repository accompanying their paper. Since this model was already trained in a supervised learning setting for the task of TCH prediction, we use a transfer learning strategy in  which we tune only the last few layers of the network. \Cref{table:xception_fine_tuning_layers} shows a small difference in training the last 2 versus the last 3 layers, so we stop there. 
For the U-Net model, we use the 
self-supervised learning for earth observation (SSL4EO-S12) weights from \cite{Wang2023}, which have been pretrained on global Sentinel-2 imagery using self-supervised learning techniques. We experiment with freezing the pretrained encoder weights and fine-tuning just the decoder versus fine-tuning the entire network for the U-Net. 

All of our models are trained to optimize a mean squared error (MSE) objective, defined pixelwise. For each model configuration and each training split, we pick the hyperparameter configuration and model weights that correspond to the best validation set loss (MSE) achieved at any point during training (up to a maximum of 200 training epochs).
For each neural network model (trained from scratch or fine-tuned), we perform a hyperparameter validation step over the initial learning rate in [0.00001, 0.0001, 0.001, 0.01] and weight decay in [0.00001, 0.0001, 0.001, 0.01]. We use the AdamW  \citep{loshchilov2017decoupled} optimizer and a learning rate scheduler that reduces the learning rate on performance plateaus by a factor of 0.1 with a patience of 10 epochs. Our neural network models are trained using the \href{https://torchgeo.readthedocs.io/en/stable/}{torchgeo} python package \citep{stewart2022torchgeo}. For the random forest models, we use the scikit-learn implementation of random forest regression \citep{scikit-learn}, and tune the maximum depth of the trees in [2, 4, 8, 16], and the number of trees in the ensemble in [50, 100, 200]. 

\subsubsection{Data preprocessing} 

To account for some anomalously high values in the LiDAR-derived TCH labels, we set any pixels greater than $30$m to no-data (NaN) values. Pixels with TCH $<0$m are set to $0$. This only occurs in a minimal percentage of the overall data; 99.95\% of the non-NaN 1m-TCH pixel values fall within 0-30m. Next, we coarsen the labels to 10m-resolution to be consistent with the resolution of Sentinel-2 imagery. When coarsening, we use the 90th percentile value of the 1m pixel values to maintain consistency with past work \citep{glad,eth}. 

We preprocess the satellite imagery input to all of our locally trained models by channelwise normalization to zero-mean, unit-variance per channel across the three Sentinel-2 tiles that span the 24 sites. When we initialize the model with weights from a previous training or pretraining procedure, we use the image preprocessing transforms used during that previous procedure, as recommended by \cite{Corley2023}.

\subsubsection{Evaluation}
We evaluate models pixelwise, at 10m resolution unless otherwise stated. When performance is reported by site, we use the models trained with the split where this site is in the test set. When performance is reported by split, we aggregate all pixels in a split to compute the average. When a single performance metric is given, this represents the average across the four splits.

When evaluating performance of the existing global TCH maps, we exclude from evaluation any pixels that contain no-data values in the global predictions (for the GLAD map, this excludes pixels corresponding to water, snow, ice, or other no-data-valued pixels). Less than 0.01\% of the non-NaN pixels in the label data have a corresponding NaN in either of the global maps for both 10m and 30m-resolution, so this has a minimal impact on the reported performance metrics.
\section{Results}
\label{sec:results}

\subsection{Evaluating existing global maps}
\label{sec:results_existing_maps} 
We first evaluate the performance of existing global TCH maps compared to our best locally trained or fine-tuned model, using both quantitative (\Cref{sec:results_existing_maps_quantitative}) and qualitative (\Cref{sec:results_existing_maps_qualitative}) assessments. We then assess the degree to which these different maps of TCH capture relative values of aboveground biomass in each site, as a way to gauge the usability of each map for additional ecological assessments in this local area (\Cref{sec:results_existing_maps_downstream}).

\subsubsection{Quantitative Performance}

\label{sec:results_existing_maps_quantitative} 
\begin{figure}[ht]
    \centering
       \includegraphics[height=3
in]{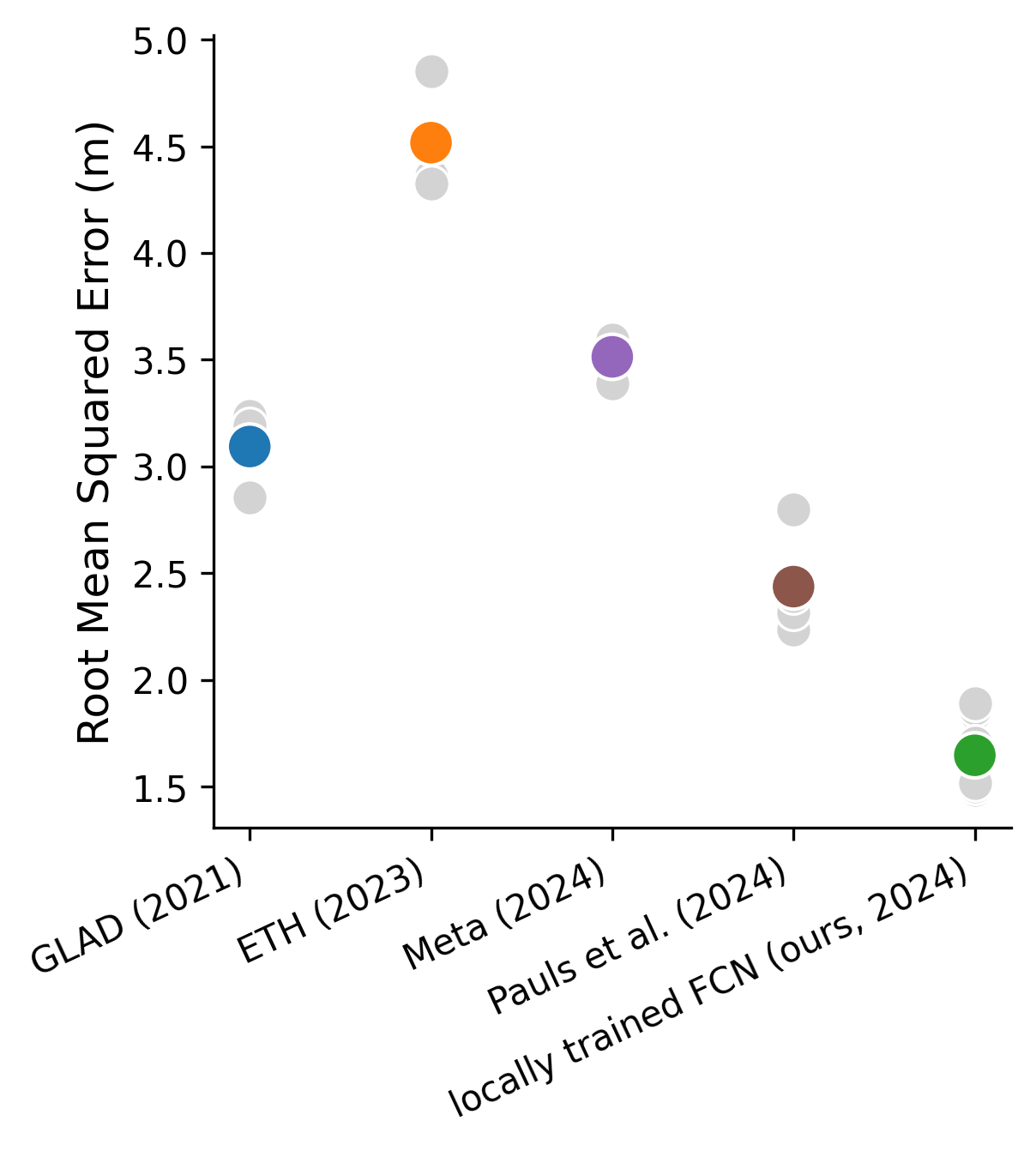}
    \caption{\textbf{A locally-trained fully convolutional network (FCN) outperforms the three existing global maps in quantitative performance}. Average performance is shown across splits (gray dots), and averaged over splits (larger, colored dots). Models on the horizontal axis are ordered by date of publication.}
    \label{fig:quantitative_map_comparison_aggregate}
\end{figure}

\begin{figure}[ht]
    \centering  \includegraphics[width=\textwidth]{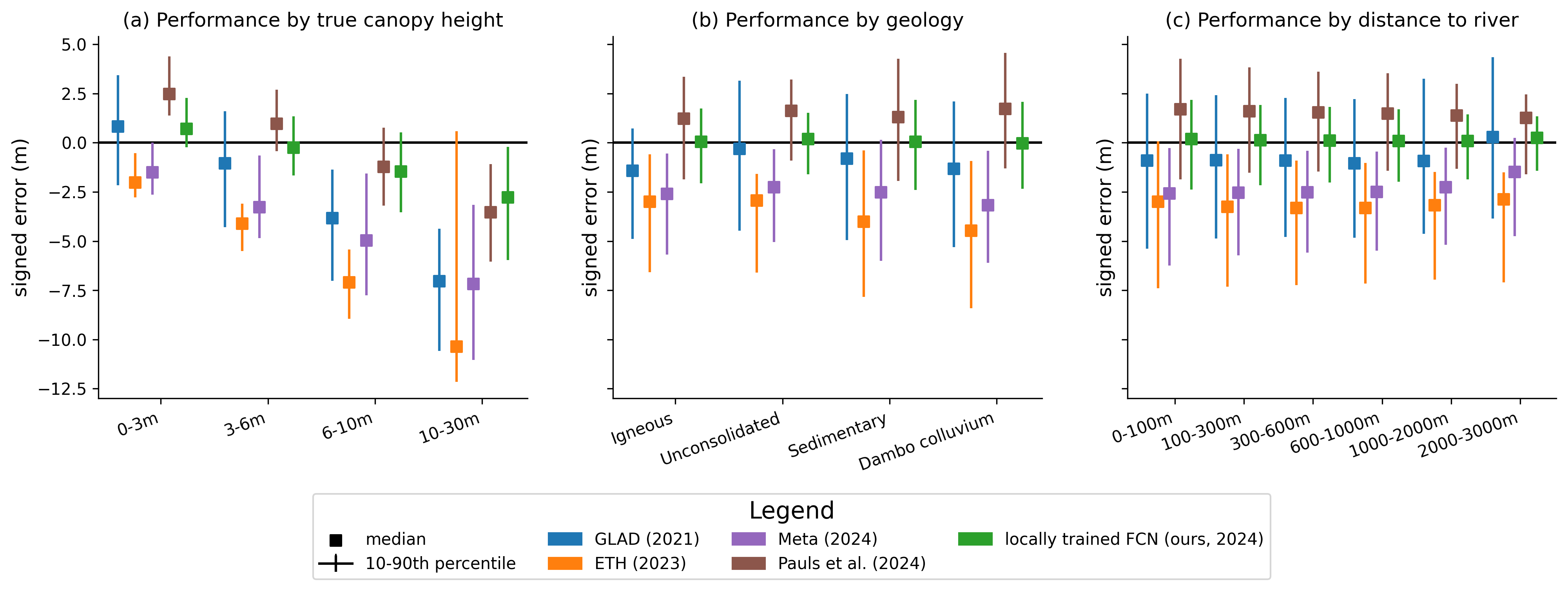}
    \caption{\textbf{A locally-trained fully convolutional network (FCN) exhibits less prediction bias than the four existing global TCH maps, across ecologically-relevant strata}: (a) binned tree canopy height, (b) geology type, and (c) binned distance to nearest river.}
\label{fig:quantitative_map_comparison_stratified}
\end{figure}

\Cref{fig:quantitative_map_comparison_aggregate} summarizes the performance of the existing global models and our best local model (with extended results in \Cref{table:fine_tuning}). The existing global maps achieve average root mean squared error (RMSE) between  2.43m and 4.51m, and for all maps the variation in performance across data splits (grey dots) is generally smaller than the variation across models. The RMSE of these maps in Karingani is significantly lower than the RMSEs of each map evaluated across the globe (3.09m vs. 9.25m  for \cite{glad}; 4.51m vs  8.62m for \cite{eth}; 2.43m vs. 4.73m  for \cite{pauls2024estimating}).\footnote{These global metrics are reported through a common evaluation procedure in \cite{pauls2024estimating}, which differ slightly from the original papers. For the purposes of our study, the relative rankings of global performance and local performance is most relevant.}
Note that for all models, the RMSE values in Karingani are lower than the global RMSEs. This indicates that the Karingani Game Reserve is not a region of especially high error for the global models, in comparison to other places around the globe.

While the global performance generally increased with each subsequent global model that has been published,
\Cref{fig:quantitative_map_comparison_aggregate} shows that 
some of the more recently published maps perform worse in Karingani than older maps. Until the map by \cite{pauls2024estimating} was released, the best existing TCH map for the Karingani region (by our quantitative metrics) was the 30m resolution map produced by \cite{glad}.  We discuss  implications of this result for understanding when and how efforts to increase global mapping performance can align with increasing local performance (our RQ3) in \Cref{sec:discussion-RQ3}.

Our locally-trained model is  superior to the existing global maps by a large margin. The predictions of our best-performing model (local FCN) achieve, on average, a 64\% lower RSME than the ETH map (4.51m → 1.64m), a 47\% lower RMSE than the GLAD map (3.09m → 1.65m), a 53\% lower RMSE than the Meta map (3.51m → 1.65m), and a 33\% lower RMSE than the \cite{pauls2024estimating} map (2.43m → 1.65m) in our study area. The gray dots in \Cref{fig:quantitative_map_comparison_aggregate} show the performance for each split individually, showing that these differences are consistent across different data splits. 


\Cref{fig:quantitative_map_comparison_stratified} disaggregates performance differences across ecologically relevant strata, including the true TCH values, geology, and distance to the nearest river corresponding to each pixel (see \Cref{appendix:stratifying-by-feature} for more details of the stratification process for these evaluations). Across these strata, our locally-trained models consistently show less bias (measured with average signed error) compared to the publicly available global models. The existing global models exhibit different biases among strata, with all models tending to under-predict pixels with higher canopy height.  All models have similarly consistent prediction bias across geology and distance to rivers, with the model of \cite{pauls2024estimating} tending to over-predict TCH, and the remaining existing models tending to under-predict TCH. Our locally trained FCN model exhibits a median of around 0 bias, despite the out-of-sample test set (\Cref{fig:problem_setting}a,b) requiring some degree of spatial generalization.

\subsubsection{Qualitative Performance}
\label{sec:results_existing_maps_qualitative} 

\begin{figure}[ht]
    \centering
    \begin{subfigure}[b]{1\textwidth}
       \includegraphics[width=1\linewidth]{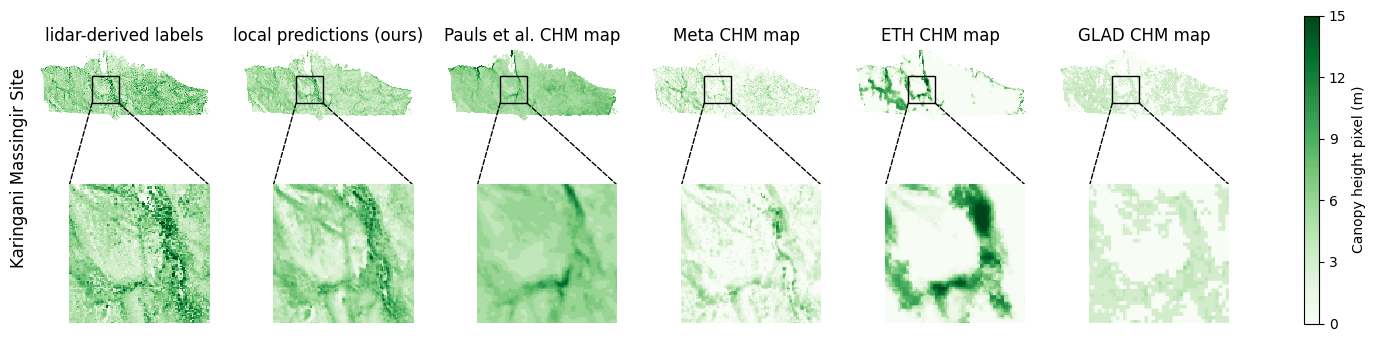}
    \end{subfigure}
    \begin{subfigure}[b]{1\textwidth}
       \includegraphics[width=1\linewidth]{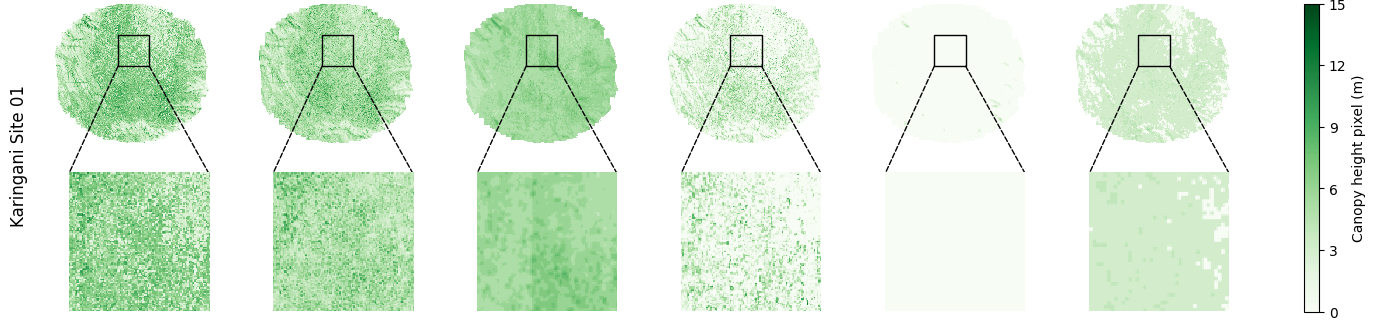}
    \end{subfigure}
    \begin{subfigure}[b]{1\textwidth}
       \includegraphics[width=1\linewidth]{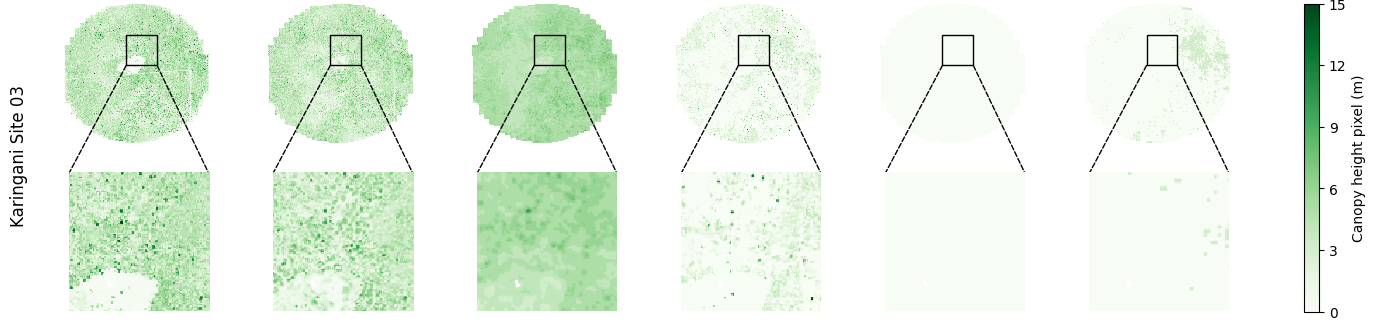}
    \end{subfigure}
    \caption{\textbf{Existing TCH maps exhibit different types of visual error structures, which are largely alleviated by our local model}. TCH labels (aggregated to 10m resolution) derived from locally collected LiDAR data, our model trained only with local labeled data (5-layer FCN with 12-channel Sentinel-2 imagery), compared with the \cite{pauls2024estimating}, Meta \citep{meta}, ETH \citep{eth}, and GLAD \citep{glad} global TCH maps at 10m resolution. Existing maps are ordered by publication date, most to least recent. 
    }
    \label{fig:qualitative_map_comparison}
\end{figure}

\Cref{fig:qualitative_map_comparison} plots our predictions and the data from the four publicly available maps for three representative sites in our study area. From visual inspection, our local predictions are clearly a better match to the LiDAR-derived TCH values than any of the existing global maps. In contrast, the GLAD map tends to under-predict TCH, and generally has more spatially smooth predictions than the label data (the original resolution of the GLAD map is 30m). The ETH map also tends to predict zeros more frequently, but sometimes captures higher canopy heights (reliably capturing higher canopy heights was a goal of the ETH study \citep{eth}). For example in the Massingir site, the ETH predictions tend toward more extreme values than the label data convey (\Cref{fig:qualitative_map_comparison}, first row). The Meta predictions capture more fine-grained textural information better than the GLAD or ETH map, but tend to underpredict TCH overall. The map from \citet{pauls2024estimating} is the closest existing map to getting the average pixel values correct, but predictions appear to biased toward the mean value, missing some of the high and low values (note also that \cite{pauls2024estimating} do not estimate any canopy heights under 3m).  
\Cref{fig:prediction_distributions} in \Cref{app:extra_results} plots the distribution of predictions and labels per pixel, aggregated across the 24 sites, confirming that these trends hold broadly across our study area.

\subsubsection{Gauging usability for downstream ecological analyses}
\label{sec:results_existing_maps_downstream}
To probe the degree to which these TCH maps could be used ``out of the box'' to assess other ecological spatial patterns related to tree canopy monitoring, we assess the degree to which each map could be locally calibrated to predict aboveground biomass (AGB). Using established allometric equations available in the ecology literature \citep{Colgan2013}, we estimate AGB (in kg) for each site using the 1m LiDAR-derived TCH values and the linear-affine formula on the average height ($H$) of pixels designated as canopy cover ($CC$).\footnote{Note that the reference AGB data are estimated from ground-truth TCH labels, not direct measurements.} Canopy cover is defined as a binary variable valued at $1$ for pixels with canopy height above a threshold of 1.5m --  \cite{Colgan2013} use TCH maps with 30m resolution, while we use 1m resolution data for our reference values, and 10m resolution data for our predicted values, except for those derived from the Meta map, which has 1m resolution).
We use the predicted values to estimate the same dependent variable $X= \textrm{average}(CC\times H)$ at each site, for each map. 
We fit a least squares regression to associate these X values with the calculated AGB values for each site, as a local calibration procedure would do. 

The $R^2$ scores (coefficient of determination) resulting from the linear-affine fits  (summarized in \Cref{fig:agb_comparison}) are 0.11 (ETH), 0.22 (GLAD), 0.52 (Meta), 0.41 (\cite{pauls2024estimating}), and 0.52 (our locally trained FCN). Overall, our locally trained FCN predictions with 10m resolution provide a signal of the relative AGB across the 24 study sites that is similar to what we would get from the 1m resolution Meta map, after a linear calibration. The map from \cite{pauls2024estimating} captures a slightly lower degree of relative variation in AGB across sites, whereas the low $R^2$ values of the fit from the ETH and GLAD maps suggest that these maps are not as well suited to estimating relative AGB across sites (at least not using this established calibration formula).

\subsection{Training local models from scratch vs. initializing with weights from global (pre)training}
\label{sec:results_finetuning} 
We next examine whether existing global models for machine learning with satellite imagery (SatML) can be a better starting place from which to fine-tune models with local data, versus training from scratch. \Cref{table:fine_tuning} compares the performance of locally fine-tuned models initialized with weights from existing global models, versus initialized randomly. 

\newcommand{\cmark}{\ding{51}}%
\newcommand{\xmark}{\ding{55}}%

\begin{table}
\centering
\resizebox{\textwidth}{!}{
\begin{tabular}{llccccccc}\toprule
 & \multicolumn{2}{c}{data used} & \multicolumn{2}{c}{RMSE (m)} &\multicolumn{2}{c}{MAE (m)} & \multicolumn{2}{c}{$R^2$} \\
\cmidrule(lr){2-3} \cmidrule(lr){4-5}\cmidrule(lr){6-7}\cmidrule(lr){8-9}
model description & local  & global  & average & std dev. & average & std dev. & average & std dev. \\\midrule
\multicolumn{9}{l}{\small \textsc{XceptionS2}} \\ 
\quad global init. & \cmark & \cmark &\underline{1.98}&0.03&\underline{1.48}&0.02&0.37&0.02\\
\quad random init. (no latlon) & \cmark & \xmark &\underline{1.94}&0.03&\underline{1.47}&0.02&\underline{0.40}&0.02\\

\quad random init. & \cmark & \xmark   &2.05&0.04&1.55&0.03&0.33&0.03\\ \hline
\multicolumn{9}{l}{\small \textsc{U-Net}} \\ 
\quad SSL init., decoder fine-tuned & \cmark & \cmark &\underline{2.05}&0.04&\underline{1.54}&0.03&\underline{0.32}&0.03\\
\quad SSL init., everything fine-tuned & \cmark & \cmark &\underline{2.08}&0.13&\underline{1.55}&0.10&\underline{0.30}&0.09 \\
\quad  random init., everything trained  & \cmark & \xmark &\underline{2.27}&0.69&\underline{1.59}&0.20&\underline{0.10}&0.69\\ \hline 
\multicolumn{9}{l}{\small \textsc{Fully Convolutional Network}} \\ 
\quad 5-layers, 128 filters & \cmark & \xmark &\textbf{\underline{1.64}}&0.01&\textbf{\underline{1.20}}&0.01&\textbf{\underline{0.57}}&$<$0.01\\ \hline
\multicolumn{9}{l}{\small \textsc{Pixelwise random forest}} \\ 
\quad parameters vary by split & \cmark & \xmark &{\underline{2.28}}&$<$0.01&{\underline{1.80}}&$<$0.01&{\underline{0.16}}&$<$0.01\\ \hline
\multicolumn{9}{l}{\small \textsc{Globally available TCH maps}} \\ 
\quad GLAD \citep{glad} & \xmark & \cmark &3.09&-&2.44&-&-0.53&-\\
\quad ETH \citep{eth} & \xmark & \cmark  &4.51&-&3.83&-&-2.26&-\\
\quad Meta \citep{meta} & \xmark & \cmark  &3.51&-&2.87&-&-0.98&-\\
\quad \cite{pauls2024estimating} & \xmark & \cmark  & \underline{2.43} &-&\underline{2.01}&-&\underline{0.05}&-\\
\bottomrule
\end{tabular}}
\caption{\textbf{Performance of fine-tuning globally pretrained models vs. training from scratch}, measured by root mean squared error (RMSE), mean absolute error (MAE), and coefficient of determination ($R^2$). Bold denotes the overall best model for each metric, underline indicates the best model of each model architecture type. The standard deviations reported are over 10 training runs with different random seeds. For ties (when average values are within 2 standard errors), we underline all tied models.  The XceptionS2 model with ``global init.'' uses weights from \cite{eth} (model 0). For all three XceptionS2 models, only the last three layers are tuned.   The U-Net models use a ResNet-18 backbone, and ``SSL init.'' denotes that model weights are initialized using the SSL4EO self-supervised learning method for Sentinel-2 satellite imagery (using data distributed across the globe) from \cite{Wang2023}. The fully convolutional network (FCN) is described throughout the paper as ``ours," as this is our best performing local-only model. }
\label{table:fine_tuning}
\end{table}

\Cref{table:fine_tuning} is organized according to two different strategies for adapting globally initialized models for local use: transferring a global supervised model for local use, and fine-tuning a task-agnostic self-supervised SatML model. We use the first strategy to adapt the XceptionS2 models -- trained by \cite{eth} for supervised learning of global TCH with GEDI LiDAR data as the original labels -- for our local use. When using this pretrained model as a starting point for local training, we deploy a transfer-learning strategy and fine-tune the last three layers of the model.  We also train randomly initialized XceptionS2 models for comparison. We introduce a variant without the location embedding input layers, in case these layers hamper performance in our site-stratified setting, which includes some degree of spatial extrapolation. The results in \Cref{table:fine_tuning} show that there are marginal differences in performance when using the pre-initialized weights versus training XceptionS2 models from scratch. Comparing within the set of XceptionS2 models we tested, the difference between the models initialized with the weights from the \cite{eth} model and the best randomly initialized XceptionS2 model is $<$0.05m  in RMSE and $<$0.02m in mean absolute error (MAE). 

Our second type of local adaption starts with a U-Net model with a ResNet-18 backbone, which has been pretrained by \cite{Wang2023} using self-supervised learning on globally-distributed Sentinel-2 satellite imagery. We experiment with fine-tuning the entire model versus fine-tuning only the decoder layers. \Cref{table:fine_tuning} shows that the largest difference in average performance between a pretrained model after fine-tuning and a model of the same architecture trained from scratch is $<$0.23m in RMSE, and $<$0.06m in MAE. We note the higher variability (across random seeds during training) in the performance of the U-Net models compared to the other model architectures.\footnote{This was in part due to model instability toward the outer edges of the image patches – we attempted to mitigate this effect during evaluation by using only the innermost $4\times4$ pixels of each $64\times64$ moving window patch when tiling our predictions as maps, though we found that the FCN and XceptionS2 models were still generally better suited to this task for our training procedure.}

Overall, the most substantial performance difference over the publicly available TCH maps (last four rows of \Cref{table:fine_tuning}) comes from using local data in any capacity. Our best locally trained model improves performance compared to the best globally produced data product by 0.79m in average RMSE and  0.81m in average MAE. We note that even our worst locally fine-tuned models from \Cref{table:fine_tuning} improve performance compared to the best globally produced data products. 
Under conditions where local data are used, the marginal differences in performance due to initializing models randomly or with pretrained weights are very small. In fact, the best overall performing model is still our small FCN model trained only with local data, with an improvement of 0.34m in average RMSE and 0.28m in average MAE compared to the best of our globally initialized, locally fine-tuned models.


\subsection{Relative importance of design decisions across modeling and data choices}
\label{sec:results_data_model_decisions}
The results detailed in \Cref{sec:results_existing_maps,sec:results_finetuning} show that models tailored for local prediction can significantly improve the performance of TCH models in local regions. 
In light of this, we investigate the extent to which several different design decisions a research team faces in designing local models affect the overall performance. One goal of these experiments is to investigate the relative value of future efforts ranging from innovating on the machine learning models, choosing which satellite data layers and sources to use as inputs to the model, and even deciding how much labeled data to collect, and where to collect them.

\begin{figure}[ht]
    \centering
    \includegraphics[width=\textwidth]{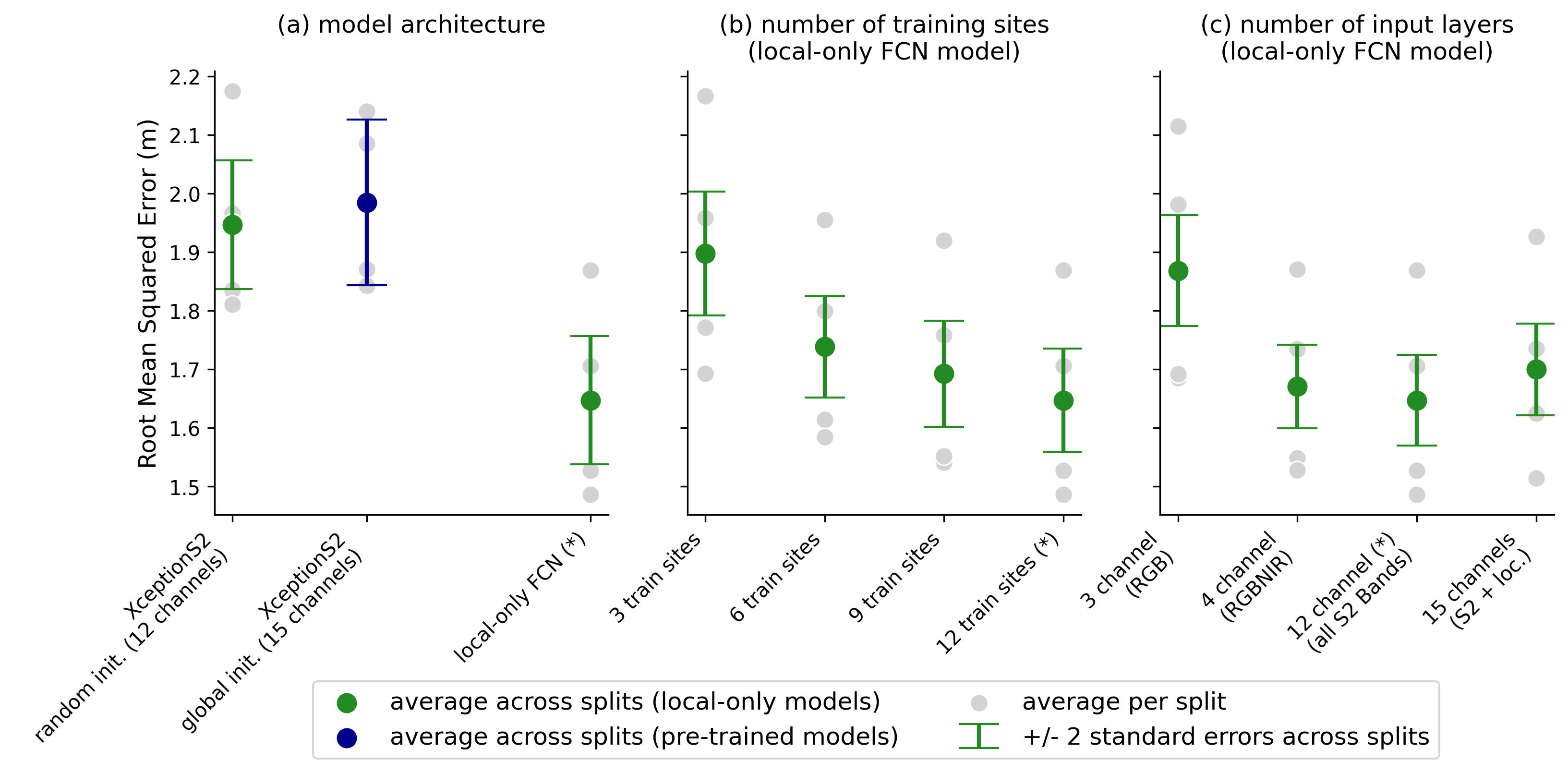}
\caption{\textbf{Different design decisions have similar magnitude of effect on local model performance}. Performance varies across: (a) different machine learning architectures, (b) amount of training data (number of training sites) available, and (c) different sets of spectral bands used as input to the model. Here S2 stands for Sentinel-2. Note that model conditions with asterisks  (*) are all the same model, for reference across panels (a-c).}
    \label{fig:data_model_choices}
\end{figure}

\Cref{fig:data_model_choices} summarizes an experiment in which we vary one of the following design decisions at a time: which machine learning model architecture is used to define the model (\Cref{fig:data_model_choices}a), which data layers are used as input to the model (\Cref{fig:data_model_choices}b), and the amount of local labeled data available for training the supervised models (\Cref{fig:data_model_choices}c). We find that each design decision has a substantial effect on overall performance. Interestingly, we find that the variation in model performance (RMSE) due to any of these factors is within roughly the same range; note the shared vertical axis range in \Cref{fig:data_model_choices}. 

When it comes to choice of model architecture, we found that simple convolutional models  can perform well at local prediction tasks (\Cref{fig:data_model_choices}a). Our best performing model is a small FCN model, with 128 filters per layer, and 5 layers, amounting to a total of 604,417 parameters (\Cref{table:num_params}), and a receptive field of only $11\times11$ pixels. This is the model we use throughout (with 12-channel Sentinel-2 inputs) as “locally-trained FCN.” The larger XceptionS2 and U-Net models had slightly worse performance, for both the randomly initialized and the globally pre-initialized variants of each model.

To understand how the availability of local data affects model performance, we take different random samples of the training sites for each of 3, 6, and 9 total sites, and compute the average model performance after training with those subsetted training sets, using our best model (locally trained FCN). To reduce the computational cost of this analysis, we fix the model hyperparameters to those used to train the local-only FCN model for each split used in the previous analyses. The total number of instances seen per epoch is kept consistent across conditions. 
The model performance resulting from using a different number of training sites is shown in \Cref{fig:data_model_choices}b. As expected, performance increases on average as the number of sites used for training increases, which is consistent across splits. The rate of returns of the average and per-split performances suggests that having even more training data would likely result in even better model performance. 

Choices made about the satellite imagery also impact performance (\Cref{fig:data_model_choices}c). For our overall best performing model architecture, performance is worst when using just the red, green, blue (RGB) spectral bands of the image, while average RMSE decreases by 10\% with the addition of a fourth near-infrared (NIR) band (\Cref{fig:data_model_choices}c). Including the remaining 8 Sentinel-2 bands has a small but positive effect on performance. Adding 3-channel location embeddings as in \cite{eth} also did not change performance greatly for our locally-trained FCN, and if anything decreased average performance. This was also the case for the locally fine-tuned XceptionS2 model (\Cref{table:fine_tuning}). 
The absolute differences in model performance across the panels in \Cref{fig:data_model_choices} contextualize the impact of different design decisions that must be considered in building a SatML model for use in local areas. Overall, the spread in performance due to these different decision factors – which model architecture to use, which data layers to use, and how much training data to use or acquire – is similar in magnitude, as indicated by the similar range of performance within each panel of \Cref{fig:data_model_choices}. The performance gap (difference in RMSE averaged across splits) due to using our second-best locally trained model (XceptionS2) versus our best (FCN), is 0.30m, while the performance gap due to using 3-channel (instead of 12-channel) imagery with the FCN is 0.22m. Likewise, the amount of training data impacts performance significantly; the difference between using 3 sites of training data versus 12 results in an average performance gap of 0.25m. 

\begin{figure}[t]
    \centering
    \includegraphics[width=\textwidth]{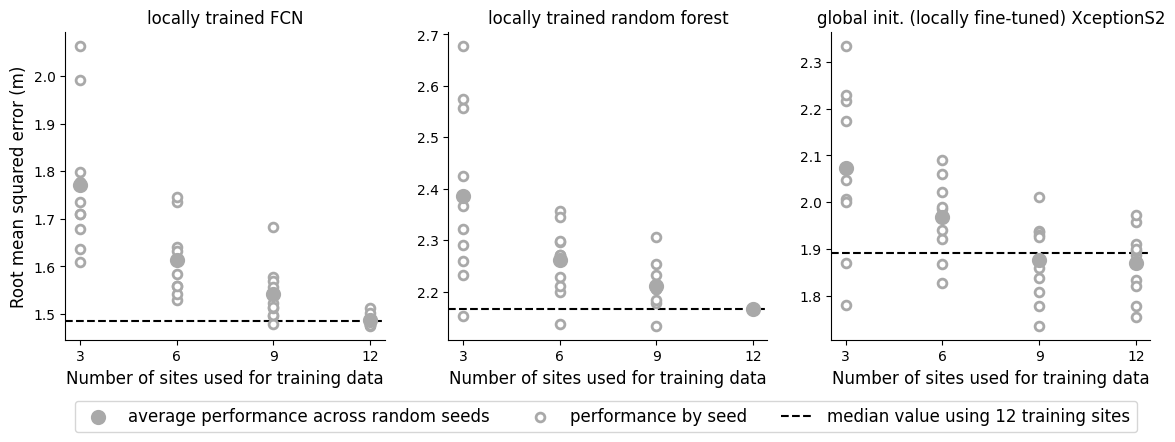}
    \caption{\textbf{Data composition and model choice interact to affect model performance} (train/val/test split 0 shown here; similar trends hold across splits --- see \Cref{fig:data_model_interactions_all_splits}). For each model, 10 random trials were run with [3,6,9,12] distinct training sites. The same 10 random subsets of training sites are used for each condition to facilitate comparison across models. }
    \label{fig:data_model_interactions}
\end{figure}

While each of these design decisions impacts local model performance in a distinct way, it is also important to understand if and how these data and modeling choices interact. 
\Cref{fig:data_model_interactions} summarizes the result of ten training runs with different subsets of 3, 6, 9 or all 12 training sites across different model architectures for a single train/val/test split (the split shown corresponds to \Cref{fig:problem_setting};
the overall trends we describe are consistent across  all four of the data splits). 
\Cref{fig:data_model_interactions} shows that the average performance generally degrades with fewer sites available for training, extending the results of \Cref{fig:data_model_choices}b across different model architectures. The magnitude of the degradation differs across models, as seen by comparing the trends in the filled-in gray dots in \Cref{fig:data_model_interactions} across each panel. 

For all models, variation in model performance across different subsets (variation in each vertical set of open gray dots) tends to increase as the number of training sites decreases (\Cref{fig:data_model_interactions}). 
For the XcpetionS2 model, a substantial number of the training runs using fewer than 12 training sites actually outperformed the median performance using 12 sites. 
Similarly, for the locally trained random forest model, the best of the 10 runs for each number of training sites was better than the best performing model using 12 sites.
Given the high variability in performance across each condition on models and number of training sites, it is not possible to conclude from this experiment alone whether the best-performing runs with subsetted data have better performance because of the data or because of the random instantiation of the model. 
Examining the random forest models, where variation is low across random seeds for the 12-training-site condition, suggests that some of the performance improvement might be due to the change in training data sites for that model.
\section{Discussion}
\label{sec:discussion}

\subsection{The value of local data for model training, fine-tuning, and evaluation}
\label{sec:discussion-RQ1}
In response to our first key research question, we find that \textbf{locally collected reference data remain pivotal for generating the best possible local maps} of tree canopy height (TCH) in our study region. 
Our first set of model comparisons (\Cref{sec:results_existing_maps}) examined the performance differences of models trained only with locally collected labels derived from UAV-LiDAR, versus using globally-available TCH maps derived from global (but sparse and noisy) GEDI data. 
While the existing global TCH maps have the benefit of being ready to use “off the shelf” without downloading any satellite imagery, training new models, or collecting local training data,  their performance in the Karangani Game Reserve area was significantly worse than what we achieved with a small FCN model trained with \emph{only} a limited amount of local, high-fidelity labeled data. This finding holds up when evaluating performance across ecologically relevant strata. Considering the potential downstream task of using TCH maps to estimate aboveground biomass, the 1m resolution Meta map \citep{meta} performed on par with our local model, while the other existing maps performed worse in this use case. This raises the importance of validating global maps for local scientific analyses, even when those maps will be calibrated for local trends.

The intermediate performance of the locally trained pixel-wise random forest (with no spatial context) in \Cref{table:fine_tuning} suggests that some, but not all, of the performance gap between models optimized for global use and models trained or fine-tuned for local use  may be due to the calibration to local label distributions.
Our second set of experiments (\Cref{sec:results_finetuning}) shows the value of local data for fine-tuning globally pre-initialized models for use in a specific region. While the best overall models we trained only used local data, we still find that models initialized with weights from global training procedures can be substantially improved by using local data for fine-tuning (note, for example, the improvement in performance of the locally fine-tuned XceptionS2 model over the published XceptionS2 global map from \cite{eth} in \Cref{table:fine_tuning}). This result underscores the potential for locally collected data to improve the regional performance of globally pretrained models, even as the usefulness of pretrained models for local use cases advances. 

Lastly, \textbf{we emphasize the importance of the locally collected data for evaluating the performance of geospatial ML models}. 
Despite an increase in high-quality local and regional TCH maps being released recently (mostly in the Global North, e.g., \cite{fogel2024opencanopy,neon}), understanding and assessing the local performance of global TCH maps across Earth's diverse continents and biomes remains difficult due to a lack of globally-distributed test sites with high quality, locally collected TCH data.  We expect that careful collection of local data will remain paramount to assessing and improving progress in geospatial machine learning for global environmental monitoring. 

\subsection{The effects of geographic-, feature-, and model-representation on  local performance}
\label{sec:discussion-RQ2}

In response to our second research question, we find that \textbf{the choices regarding (i) what model architecture to use, (ii) what data layers to use as input, and (iii) where label data are collected can all have a large impact on local performance}. 
Overall, there are substantial differences in performance across the model architectures we trained and fine-tuned with local data (\Cref{table:fine_tuning}). 
Notably, all of the models we used here have been proposed and used in prior works, and are not designed specifically for our local TCH modeling task. Thus, it is possible that local performance in areas like the Karingani Game Reserve could be further improved with modeling innovations -- either by designing models specifically for local training, or by designing global models to be more suitable for downstream adaptation in smaller local areas.

Data choices similarly had a large impact on local performance. The availability of local label data was crucial for success in training and fine-tuning models for local use in the Karingani Game Reserve. 
The choice of which spectral bands and geographic data layers to use as input also had a significant impact on performance.
In particular, the inclusion of additional spectral bands (beyond 3-channel RGB) decreased the average error of our best model by 10\%. For future researchers, decisions about which imagery to use should be based not only the spectral bands available, but also on the spatial and temporal resolution of the available imagery, which can be pivotal to model success. We expect that the difference in performance due to including all spectral bands of Sentinel-2 imagery is a conservative estimate of the performance differences that would arise from using different satellite products (e.g. Landsat or Maxar satellite imagery, which have different resolutions).

From our experiments in \Cref{sec:results_data_model_decisions}, we find that \textbf{the performance differences due to data design decisions were similar to the performance differences due to model architecture decisions.}
The absolute difference in local performance (measured against our best model, which used all training sites and all Sentinel-2 input layers) is strikingly similar across the three different types of variations we tested: (i) using a model architecture from recent TCH-mapping research efforts, (ii) using only RGB spectral bands from the satellite imagery input, or (iii) using only 3 randomly selected training sites instead of 12. 
From these findings, we conclude that the quantity and composition of the available local training data sites affected model performance with an effect comparable to what we might expect from innovations in model architectures (as roughly reflected by the gap in performance between our best and second-best models in \Cref{fig:data_model_choices}a).

\subsection{Identifying points of conflict and synergy in local and global mapping}
\label{sec:discussion-RQ3}
Our third research question concerns identifying points of potential disparity or cohesion between the goals of local and global modeling in geospatial machine learning. 
While global maps will clearly exhibit different errors in different locations (average performance in certain regions will  differ from average global performance), we might hope that the existence of global models would make it easier to build good models for local use. However, the results of our case study underscore that \textbf{currently, globally pretrained models are not necessarily the ``right'' starting point for developing good local models.}

After extensive experimentation, existing globally trained models did not factor in to the best local modeling solution we found in our case study.  At the same time, we expect that our local models would perform terribly across the globe, due to domain shifts and lack of global diversity in our training data. This discrepancy suggests that there are at least some points of conflict between local and global mapping endeavors.   
We caveat that since our case study concerns a single prediction task and study area (which allowed us to focus on the nuances of a specific local downstream use case), our discussion of RQ3 -- as it pertains to geospatial modeling across general tasks -- is more speculative in nature than our discussion of our first two research questions. 
Interpreting the differences in performance from past global TCH mapping efforts in the context of a broader set of geospatial ML studies, we can still generate insights as to which modeling choices that influenced the differences in local performance in the Karingani Game Reserve are likely to generalize more broadly across different geospatial machine learning tasks.

Our results suggest that \textbf{the best decisions about which data to use during training may differ for local and global use cases}, both in terms of which regions to include training data from, and which input layers to use. 
Regarding the \emph{regions that training data are drawn from}, we hypothesize that the relatively lasting success of the GLAD map \citep{glad} is due in part to the fact that is actually an accumulation of regional models applied across the globe. This hypothesis is consistent with results of previous work \citep{Healey2020,Tsao2023}, which showed that there may be an optimal radius of training data that is large enough to constitute enough volume and diversity of training data, yet small enough to generate specialized, accurate local models. 
Regarding \emph{input layers}, \cite{eth} included positional encoding layers (derivatives of latitude and longitude) to achieve good global performance. However, for our local-mapping task, including positional encoding layers did not help with local fine-tuning performance, and possibly were a detriment to local performance (\Cref{table:fine_tuning} and \Cref{fig:data_model_choices}). 
One possible reason for this result is that while direct positional encodings can help with spatial interpolation (e.g. filling in the gaps in sparse but global GEDI data), they may not be necessary over smaller spatial extents, or may even hinder performance in spatial extrapolation settings (e.g. extending the spatial range of local training data). Similar concerns have also been raised in other geospatial machine learning domains \citep{meyer2019importance,lilja2024localization}. 

The differences in the performance of existing global TCH maps is consistent with the hypothesis that \textbf{modeling efforts that account for and correct local error structures in global labels may be an important step to synergizing local and global supervised learning efforts}. 
Only one of the three most recent global TCH maps \citep{pauls2024estimating} improved upon the performance of the original 2021 GLAD global map, when evaluated in Karingani. In their study, \cite{pauls2024estimating} explicitly aimed to remedy gaps in local performance of global TCH models by accounting for error structures in the global GEDI data they used for training. Specifically, they accounted for localization error due to uncertainty in the position of each GEDI data point, and they removed observations from high slopes (which make the satellite-based LiDAR measurements especially noisy) during training.
The general strategy of accounting for and correcting error structures can be applied in many geospatial ML domains where some form of global labels may be available, but these labels likely come with localization noise, label noise, or statistical biases 
\citep{kuffer2022missing,gevaert2024auditing,Rolf2024}. 

\section{General takeaways for geospatial machine learning}
\label{sec:takeaways}
In addition to addressing our three key research questions as discussed in \Cref{sec:discussion}, the results from our case study provide three concrete takeaways with implications for future efforts in geospatial machine learning at different spatial scales.

\textbf{Takeaway 1: The best global models and the best local models may differ in model design and training strategy.} 
Our results provide evidence that geospatial machine learning models that capture global variation well -- as all four existing global maps of tree canopy height do -- might not sufficiently capture local variation in the target variable of interest without significant local calibration or fine-tuning. 
Our results reinforce scholarship that found that SatML models trained in large areas often break down when evaluated in local regions, whereas locally-trained models have been evidenced to do much better. Specifically, the breakdown of global maps in local regions has been shown across domains including crop land classification \citep{Kerner2024}, population density estimation \citep{kuffer2022missing}, building detection \citep{gevaert2024auditing}, and poverty mapping \citep{Aiken2023}. 

Our best model for mapping TCH in the Karingani Game Reserve was a relatively small model trained only with local data (LiDAR-derived labels and Sentinel-2 imagery). This suggests that when training local models or pretraining global models for local use, it might be advantageous to use smaller models. At the same time,  ultra-small models, such as a pixelwise random forest, obtained reasonable performance but performed worse than models that incorporate spatial context (e.g. texture) in images. Additionally, we found that design decisions aimed to enhance global modeling, such as including direct location embeddings as input to the model, may not be necessary for local models, and might actually hinder  performance in a local mapping setting.

Local and global mapping efforts are both crucial to informing scientific knowledge about the Earth and its environment. 
At the same, both efforts have significant costs, and it should not be assumed that global models will be sufficient for capturing local variation, without modification. 
This brings us to our next major takeaway, concerning the creation of ML models for environmental monitoring for use across both global and local scales.

\textbf{Takeaway 2: If geospatial machine learning models need to jointly satisfy global and local monitoring use cases, then novel research innovations may be needed.}
It is tempting to expect that working to improve global performance for geospatial ML models will implicitly result in models that are more useful for local modeling tasks as well.
However, results from \Cref{sec:results_finetuning} suggest that at present, the tree canopy height models designed for global prediction do not necessarily provide a better model initialization for our local mapping task in the Karingani Game Reserve. 
Given the substantial costs of training global models, and the additional substantial cost of collecting local data to calibrate global models (or train entirely local models), it is important to find synergy between the efforts made for global and local modeling with geospatial ML.   

Our findings suggest that fundamental innovations in this general direction of research may be needed to achieve the practical goal of building global models suitable for downstream use in local settings.
Straightforward fine-tuning or calibration of global models did not garner the best local performance for tree canopy height mapping in the Karingani Game Reserve region. However, there are still several areas of opportunity for developing synergistic modeling solutions where models trained with global data may help with downstream local tasks. For example, model distillation methods \citep{polino2018model,vemulapalli2023label} may hold promise for improved fine-tuning of large global models for local tasks. Another avenue to design models for local use cases might explicitly design the pretraining strategy for different local regions, similar to how meta-learning has been used to initialize geospatial ML models used for different prediction task domains \citep{Rubwurm2024}. 
Our results have several implications for how to  build geospatial ML models with global data, especially if the intent is to fine-tune these models for local contexts.

\textbf{Takeaway 3: Data-centric approaches may be a promising avenue for jointly achieving local and global performance with geospatial machine learning.}
Several different aspects of our results indicate that data-centric innovations could be key to expanding the joint frontiers of local and global mapping efforts. 
As discussed in \Cref{sec:discussion-RQ3},
the best of the existing global models in terms of both global and local performance mostly focuses on curating and preprocessing the training data \citep{pauls2024estimating}, while the next best model (the GLAD model of \cite{glad}) is actually an accumulation of regional models.
Additionally, our results in \Cref{sec:results_data_model_decisions} indicate that even a very coarse and straightforward experimentation with different data conditions and processing strategies could lead to substantial increases in local performance. 


Many different approaches to data-centric geospatial machine learning exist and are being developed -- see the rich discussion in \cite{roscher2024better}. 
Adding to this discussion, our results underscore the importance of studying how best practices regarding data and model decisions in SatML might differ for local and global mapping paradigms. 
Better guidelines are needed for how best to collect ground-referenced data, which data source(s) to use for which tasks, and how best to preprocess global and local data for training geospatial machine learning models  -- across a variety of application domains and prediction settings ranging from global to hyperlocal.
\section{Conclusion}
In this paper, we studied perspectives of local and global mapping with machine learning and satellite imagery, rooted in a case study of tree canopy height mapping in the ecologically diverse Karingani Game Reserve in Mozambique. 
Tree canopy height prediction is a setting in which both global and local maps are pivotal to inform environmental and ecological science and policy.
Global maps are crucial for understanding planetary ecosystem and tree composition, while local maps are crucial for scientific studies in landscape ecology and informing locally-enacted policies. 
The same argument -- that both global and local maps are fundamentally needed -- can be made for virtually any environmental or ecological variable of interest, from land cover classification to air quality estimation. 
Consequently, we need geospatial machine learning models that can resolve both large-scale global variation and fine-grained local variation in key environmental variables. 

Our study focused on the task of tree canopy height mapping in Karingani Game Reserve, which has not been previously used to train or evaluate geospatial ML models. As such, these data enabled us to ground our study in considerations researchers and modelers face when applying geospatial machine learning for ecological and environmental modeling to a new area. As discussed in \Cref{sec:discussion}, we expect that our main findings will generalize to different local areas and to different prediction tasks due to evidence of similar local-global map discrepancies in other geospatial prediction domains. That said, we expect that the exact numbers and nuanced insights may differ from setting to setting. It would be informative to repeat our study across many local areas and similar prediction tasks in future work (made possible with a growing number of public datasets representing mapping tasks related to trees \citep{Weinstein2021,Reiersen2022,Ouaknine2023,Ahlswede2023,Bountos2024}).

Our results indicate that, at present, global and local mapping may constitute distinct modeling goals for geospatial machine learning. This hypothesis is  supported by previous findings across tasks including cropland mapping \citep{Kerner2024}, poverty prediction \citep{Aiken2023}, population density estimation \citep{kuffer2022missing}, and building detection \citep{gevaert2024auditing}. Even the best global models for tree canopy height mapping, which represent impressive efforts in data processing and model architecture \citep{glad,eth,meta,pauls2024estimating} are outperformed by simple fully convolutional networks trained on roughly 250 square kilometers of carefully collected highly accurate label data paired with publicly available Sentinel-2 satellite imagery. Moreover, we find that some of the design decisions that past work found important for global modeling were either unnecessary or potentially harmful to local performance. 
%

Looking forward, there is a great opportunity to develop models that can flexibly interpolate between different prediction extents, ranging from global to hyperlocal. 
Our study exposes a need for future research directions that prioritize the performance of geospatial machine learning models in local areas -- for example models pretrained on global satellite imagery to be efficiently adapted for local use -- alongside global performance. Distinguishing between the goals of local and global modeling is a first step to establishing best practices for model and design decisions for applications of geospatial machine learning for environmental monitoring. It is also an important step toward developing learning paradigms that enhance the synergy between the two efforts, for example, building models that flexibly leverage both local and global data depending on data availability and the prediction region at hand. In addition to synergistic local-global modeling efforts, an emphasis on the quality of local data for model evaluation and calibration remains critical for local, global, and synergistic model developments.

\section*{Acknowledgements}{The authors thank Nico Lang for answering questions that helped us implement the XceptionS2 models from \cite{eth} and Konstantin Klemmer for feedback on an earlier version of this manuscript. Karingani Game Reserve is thanked for permission to collect LiDAR data, as well as for significant logistical support during data collection, with particular thanks to Ellery Worth for on the ground support. Pete Goodman is thanked for providing spatial data on rivers and geology. Tom Lautenbach and Michael Voysey are thanked for helping with data collection, and Jenia Singh and Peter Boucher for LiDAR data processing. Karingani Holdings are acknowledged for funding the LiDAR data collection.
\\\\
E.R. was supported by postdoctoral fellowships from the Harvard Data Science Initiative and the Harvard Center for Research on Computation and Society.
L.G. was supported by the National Science Foundation Graduate Research Fellowship under Grant No. DGE2140743. }
\newpage

\bibliographystyle{plainnat}
\bibliography{localmapping}

\begin{thebibliography}{52}
\providecommand{\natexlab}[1]{#1}
\providecommand{\url}[1]{\texttt{#1}}
\expandafter\ifx\csname urlstyle\endcsname\relax
  \providecommand{\doi}[1]{doi: #1}\else
  \providecommand{\doi}{doi: \begingroup \urlstyle{rm}\Url}\fi

\bibitem[Ahlswede et~al.(2023)Ahlswede, Schulz, Gava, Helber, Bischke, F\"orster, Arias, Hees, Demir, and Kleinschmit]{Ahlswede2023}
S.~Ahlswede, C.~Schulz, C.~Gava, P.~Helber, B.~Bischke, M.~F\"orster, F.~Arias, J.~Hees, B.~Demir, and B.~Kleinschmit.
\newblock {TreeSatAI} benchmark archive: a multi-sensor, multi-label dataset for tree species classification in remote sensing.
\newblock \emph{Earth System Science Data}, 15\penalty0 (2):\penalty0 681--695, 2023.
\newblock \doi{10.5194/essd-15-681-2023}.

\bibitem[Aiken et~al.(2023)Aiken, Rolf, and Blumenstock]{Aiken2023}
Emily Aiken, Esther Rolf, and Joshua Blumenstock.
\newblock Fairness and representation in satellite-based poverty maps: Evidence of urban-rural disparities and their impacts on downstream policy.
\newblock In \emph{Proceedings of the Thirty-Second International Joint Conference on Artificial Intelligence}, 2023.
\newblock ISBN 978-1-956792-03-4.
\newblock \doi{10.24963/ijcai.2023/653}.

\bibitem[Aponte et~al.(2020)Aponte, Kasel, Nitschke, Tanase, Vickers, Parker, Fedrigo, Kohout, Ruiz-Benito, Zavala, et~al.]{aponte2020structural}
Cristina Aponte, Sabine Kasel, Craig~R Nitschke, Mihai~A Tanase, Helen Vickers, Linda Parker, Melissa Fedrigo, Michele Kohout, Paloma Ruiz-Benito, Miguel~A Zavala, et~al.
\newblock Structural diversity underpins carbon storage in {A}ustralian temperate forests.
\newblock \emph{Global Ecology and Biogeography}, 29\penalty0 (5):\penalty0 789--802, 2020.

\bibitem[Astola et~al.(2021)Astola, Seitsonen, Halme, Molinier, and Lönnqvist]{Astola2021}
Heikki Astola, Lauri Seitsonen, Eelis Halme, Matthieu Molinier, and Anne Lönnqvist.
\newblock Deep neural networks with transfer learning for forest variable estimation using {Sentinel-2} imagery in boreal forest.
\newblock \emph{Remote Sensing}, 13\penalty0 (12), 2021.
\newblock ISSN 2072-4292.
\newblock \doi{10.3390/rs13122392}.

\bibitem[Boucher et~al.(2023)Boucher, Hockridge, Singh, and Davies]{boucher2023flying}
Peter~B Boucher, Evan~G Hockridge, Jenia Singh, and Andrew~B Davies.
\newblock Flying high: Sampling savanna vegetation with {UAV}-lidar.
\newblock \emph{Methods in Ecology and Evolution}, 14\penalty0 (7):\penalty0 1668--1686, 2023.

\bibitem[Bountos et~al.(2023)Bountos, Ouaknine, and Rolnick]{Bountos2024}
Nikolaos~Ioannis Bountos, Arthur Ouaknine, and David Rolnick.
\newblock {FoMo-Bench}: a multi-modal, multi-scale and multi-task forest monitoring benchmark for remote sensing foundation models.
\newblock \emph{arXiv preprint arXiv:2312.10114}, 2023.

\bibitem[Brown et~al.(2022)Brown, Brumby, Guzder-Williams, Birch, Hyde, Mazzariello, Czerwinski, Pasquarella, Haertel, Ilyushchenko, Schwehr, Weisse, Stolle, Hanson, Guinan, Moore, and Tait]{Brown2022}
Christopher~F. Brown, Steven~P. Brumby, Brookie Guzder-Williams, Tanya Birch, Samantha~Brooks Hyde, Joseph Mazzariello, Wanda Czerwinski, Valerie~J. Pasquarella, Robert Haertel, Simon Ilyushchenko, Kurt Schwehr, Mikaela Weisse, Fred Stolle, Craig Hanson, Oliver Guinan, Rebecca Moore, and Alexander~M. Tait.
\newblock Dynamic {W}orld, near real-time global 10m land use land cover mapping.
\newblock \emph{Scientific Data}, 9\penalty0 (1), June 2022.
\newblock ISSN 2052-4463.
\newblock \doi{10.1038/s41597-022-01307-4}.

\bibitem[Chi et~al.(2022)Chi, Fang, Chatterjee, and Blumenstock]{Chi2022}
Guanghua Chi, Han Fang, Sourav Chatterjee, and Joshua~E. Blumenstock.
\newblock Microestimates of wealth for all low- and middle-income countries.
\newblock \emph{Proceedings of the National Academy of Sciences}, 119\penalty0 (3), January 2022.
\newblock ISSN 1091-6490.
\newblock \doi{10.1073/pnas.2113658119}.
\newblock URL \url{http://dx.doi.org/10.1073/pnas.2113658119}.

\bibitem[Colgan et~al.(2013)Colgan, Asner, and Swemmer]{Colgan2013}
Matthew~S. Colgan, Gregory~P. Asner, and Tony Swemmer.
\newblock Harvesting tree biomass at the stand level to assess the accuracy of field and airborne biomass estimation in savannas.
\newblock \emph{Ecological Applications}, 23\penalty0 (5):\penalty0 1170--1184, 2013.
\newblock \doi{https://doi.org/10.1890/12-0922.1}.

\bibitem[Corley et~al.(2024)Corley, Robinson, Dodhia, Ferres, and Najafirad]{Corley2023}
Isaac Corley, Caleb Robinson, Rahul Dodhia, Juan M~Lavista Ferres, and Peyman Najafirad.
\newblock Revisiting pre-trained remote sensing model benchmarks: resizing and normalization matters.
\newblock In \emph{Proceedings of the IEEE/CVF Conference on Computer Vision and Pattern Recognition}, pages 3162--3172, 2024.

\bibitem[Coverdale and Davies(2023)]{coverdale2023unravelling}
Tyler~C Coverdale and Andrew~B Davies.
\newblock Unravelling the relationship between plant diversity and vegetation structural complexity: A review and theoretical framework.
\newblock \emph{Journal of Ecology}, 111\penalty0 (7):\penalty0 1378--1395, 2023.

\bibitem[Davies and Asner(2014)]{davies2014advances}
Andrew~B Davies and Gregory~P Asner.
\newblock Advances in animal ecology from {3D-LiDAR} ecosystem mapping.
\newblock \emph{Trends in ecology \& evolution}, 29\penalty0 (12):\penalty0 681--691, 2014.

\bibitem[Davies et~al.(2016)Davies, Tambling, Kerley, and Asner]{davies2016effects}
Andrew~B Davies, Craig~J Tambling, Graham~IH Kerley, and Gregory~P Asner.
\newblock Effects of vegetation structure on the location of lion kill sites in {A}frican thicket.
\newblock \emph{PloS one}, 11\penalty0 (2):\penalty0 e0149098, 2016.

\bibitem[Fogel et~al.(2024)Fogel, Perron, Besic, Saint-Andr{\'e}, Pellissier-Tanon, Schwartz, Boudras, Fayad, d'Aspremont, Landrieu, et~al.]{fogel2024opencanopy}
Fajwel Fogel, Yohann Perron, Nikola Besic, Laurent Saint-Andr{\'e}, Agn{\`e}s Pellissier-Tanon, Martin Schwartz, Thomas Boudras, Ibrahim Fayad, Alexandre d'Aspremont, Loic Landrieu, et~al.
\newblock {Open-Canopy}: A country-scale benchmark for canopy height estimation at very high resolution.
\newblock \emph{arXiv preprint arXiv:2407.09392}, 2024.

\bibitem[Gevaert et~al.(2024)Gevaert, Buunk, and Van Den~Homberg]{gevaert2024auditing}
Caroline~M Gevaert, Thomas Buunk, and Marc~JC Van Den~Homberg.
\newblock Auditing geospatial datasets for biases: using global building datasets for disaster risk management.
\newblock \emph{IEEE Journal of Selected Topics in Applied Earth Observations and Remote Sensing}, 2024.

\bibitem[Healey et~al.(2020)Healey, Yang, Gorelick, and Ilyushchenko]{Healey2020}
Sean~P. Healey, Zhiqiang Yang, Noel Gorelick, and Simon Ilyushchenko.
\newblock Highly local model calibration with a new {GEDI LiDAR} asset on {Google Earth Engine} reduces {L}andsat forest height signal saturation.
\newblock \emph{Remote Sensing}, 12\penalty0 (17), 2020.
\newblock ISSN 2072-4292.
\newblock \doi{10.3390/rs12172840}.

\bibitem[Huang et~al.(2008)Huang, Yang, King, and Lyu]{Huang2008}
Kaizhu Huang, Haiqin Yang, Irwin King, and Michael Lyu.
\newblock \emph{{Machine Learning: Modeling Data Locally and Globally}}.
\newblock Zhejiang University Press and Springer Berlin Heidelberg, Hangzhou and Berlin, 2008.

\bibitem[Kerner et~al.(2024)Kerner, Nakalembe, Yang, Zvonkov, McWeeny, Tseng, and Becker-Reshef]{Kerner2024}
Hannah Kerner, Catherine Nakalembe, Adam Yang, Ivan Zvonkov, Ryan McWeeny, Gabriel Tseng, and Inbal Becker-Reshef.
\newblock {How accurate are existing land cover maps for agriculture in Sub-Saharan Africa?}
\newblock \emph{Scientific Data}, 11\penalty0 (486), 2024.

\bibitem[Kuffer et~al.(2022)Kuffer, Owusu, Oliveira, Sliuzas, and van Rijn]{kuffer2022missing}
Monika Kuffer, Maxwell Owusu, Lorraine Oliveira, Richard Sliuzas, and Frank van Rijn.
\newblock The missing millions in maps: exploring causes of uncertainties in global gridded population datasets.
\newblock \emph{ISPRS International Journal of Geo-Information}, 11\penalty0 (7):\penalty0 403, 2022.

\bibitem[Lang et~al.(2023)Lang, Jetz, Schindler, and Wegner]{eth}
Nico Lang, Walter Jetz, Konrad Schindler, and Jan~Dirk Wegner.
\newblock {A high-resolution canopy height model of the Earth}.
\newblock \emph{Nature Ecology \& Evolution}, 7, 2023.

\bibitem[Li et~al.(2023)Li, Wessels, Armston, Hancock, Mathieu, Main, Naidoo, Erasmus, and Scholes]{Li2023}
Xiaoxuan Li, Konrad Wessels, John Armston, Steven Hancock, Renaud Mathieu, Russell Main, Laven Naidoo, Barend Erasmus, and Robert Scholes.
\newblock First validation of {GEDI} canopy heights in {A}frican savannas.
\newblock \emph{Remote Sensing of Environment}, 285:\penalty0 113402, 2023.
\newblock ISSN 0034-4257.
\newblock \doi{https://doi.org/10.1016/j.rse.2022.113402}.

\bibitem[Lilja et~al.(2024)Lilja, Fu, Stenborg, and Hammarstrand]{lilja2024localization}
Adam Lilja, Junsheng Fu, Erik Stenborg, and Lars Hammarstrand.
\newblock Localization is all you evaluate: Data leakage in online mapping datasets and how to fix it.
\newblock In \emph{Proceedings of the IEEE/CVF Conference on Computer Vision and Pattern Recognition}, pages 22150--22159, 2024.

\bibitem[Loarie et~al.(2013)Loarie, Tambling, and Asner]{loarie2013lion}
Scott~R Loarie, Craig~J Tambling, and Gregory~P Asner.
\newblock Lion hunting behaviour and vegetation structure in an {A}frican savanna.
\newblock \emph{Animal Behaviour}, 85\penalty0 (5):\penalty0 899--906, 2013.

\bibitem[Loshchilov and Hutter(2019)]{loshchilov2017decoupled}
Ilya Loshchilov and Frank Hutter.
\newblock Decoupled weight decay regularization.
\newblock \emph{International Conference on Learning Representations}, 2019.

\bibitem[Meyer et~al.(2019)Meyer, Reudenbach, W{\"o}llauer, and Nauss]{meyer2019importance}
Hanna Meyer, Christoph Reudenbach, Stephan W{\"o}llauer, and Thomas Nauss.
\newblock Importance of spatial predictor variable selection in machine learning applications--moving from data reproduction to spatial prediction.
\newblock \emph{Ecological Modelling}, 411:\penalty0 108815, 2019.

\bibitem[{National Ecological Observatory Network (NEON)}(2024)]{neon}
{National Ecological Observatory Network (NEON)}.
\newblock Ecosystem structure (dp3.30015.001), 2024.
\newblock URL \url{https://data.neonscience.org/data-products/DP3.30015.001/RELEASE-2024}.

\bibitem[Noul{\`e}koun et~al.(2021)Noul{\`e}koun, Birhane, Mensah, Kassa, Berhe, Gebremichael, Adem, Seyoum, Mengistu, Lemma, et~al.]{noulekoun2021structural}
Florent Noul{\`e}koun, Emiru Birhane, Sylvanus Mensah, Habtemariam Kassa, Alemayehu Berhe, Zefere~Mulaw Gebremichael, Nuru~Mohammed Adem, Yigremachew Seyoum, Tefera Mengistu, Bekele Lemma, et~al.
\newblock Structural diversity consistently mediates species richness effects on aboveground carbon along altitudinal gradients in northern {E}thiopian grazing exclosures.
\newblock \emph{Science of the Total Environment}, 776:\penalty0 145838, 2021.

\bibitem[Nyandwi et~al.(2024)Nyandwi, Gerke, and Achanccaray]{nyandwi2024local}
Emmanuel Nyandwi, Markus Gerke, and Pedro Achanccaray.
\newblock Local evaluation of large-scale remote sensing machine learning-generated building and road dataset: The case of {R}wanda.
\newblock \emph{PFG--Journal of Photogrammetry, Remote Sensing and Geoinformation Science}, pages 1--18, 2024.

\bibitem[Ouaknine et~al.(2023)Ouaknine, Kattenborn, Lalibert{\'e}, and Rolnick]{Ouaknine2023}
Arthur Ouaknine, Teja Kattenborn, Etienne Lalibert{\'e}, and David Rolnick.
\newblock {OpenForest}: A data catalogue for machine learning in forest monitoring.
\newblock \emph{arXiv preprint arXiv:2311.00277}, 2023.

\bibitem[Paolo et~al.(2024)Paolo, Kroodsma, Raynor, Hochberg, Davis, Cleary, Marsaglia, Orofino, Thomas, and Halpin]{Paolo2024}
Fernando~S. Paolo, David Kroodsma, Jennifer Raynor, Tim Hochberg, Pete Davis, Jesse Cleary, Luca Marsaglia, Sara Orofino, Christian Thomas, and Patrick Halpin.
\newblock Satellite mapping reveals extensive industrial activity at sea.
\newblock \emph{Nature}, 625\penalty0 (7993):\penalty0 85–91, January 2024.
\newblock ISSN 1476-4687.
\newblock \doi{10.1038/s41586-023-06825-8}.

\bibitem[Pauls et~al.(2024)Pauls, Zimmer, Kelly, Schwartz, Saatchi, Ciais, Pokutta, Brandt, and Gieseke]{pauls2024estimating}
Jan Pauls, Max Zimmer, Una~M Kelly, Martin Schwartz, Sassan Saatchi, Philippe Ciais, Sebastian Pokutta, Martin Brandt, and Fabian Gieseke.
\newblock Estimating canopy height at scale.
\newblock \emph{International Conference of Machine Learning}, 2024.

\bibitem[Pedregosa et~al.(2011)Pedregosa, Varoquaux, Gramfort, Michel, Thirion, Grisel, Blondel, Prettenhofer, Weiss, Dubourg, Vanderplas, Passos, Cournapeau, Brucher, Perrot, and Duchesnay]{scikit-learn}
F.~Pedregosa, G.~Varoquaux, A.~Gramfort, V.~Michel, B.~Thirion, O.~Grisel, M.~Blondel, P.~Prettenhofer, R.~Weiss, V.~Dubourg, J.~Vanderplas, A.~Passos, D.~Cournapeau, M.~Brucher, M.~Perrot, and E.~Duchesnay.
\newblock Scikit-learn: Machine learning in {P}ython.
\newblock \emph{Journal of Machine Learning Research}, 12:\penalty0 2825--2830, 2011.

\bibitem[Polino et~al.(2018)Polino, Pascanu, and Alistarh]{polino2018model}
Antonio Polino, Razvan Pascanu, and Dan Alistarh.
\newblock Model compression via distillation and quantization.
\newblock In \emph{International Conference on Learning Representations}, 2018.

\bibitem[Potapov et~al.(2021)Potapov, Li, Hernandez-Serna, Tyukavina, Hansen, Kommareddy, Pickens, Turubanova, Tang, Silva, Armston, Dubayah, Blair, and Hofton]{glad}
Peter Potapov, Xinyuan Li, Andres Hernandez-Serna, Alexandra Tyukavina, Matthew~C. Hansen, Anil Kommareddy, Amy Pickens, Svetlana Turubanova, Hao Tang, Carlos~Edibaldo Silva, John Armston, Ralph Dubayah, J.~Bryan Blair, and Michelle Hofton.
\newblock Mapping global forest canopy height through integration of {GEDI} and {L}andsat data.
\newblock \emph{Remote Sensing of Environment}, 253:\penalty0 112165, 2021.
\newblock ISSN 0034-4257.
\newblock \doi{https://doi.org/10.1016/j.rse.2020.112165}.

\bibitem[Reiersen et~al.(2022)Reiersen, Dao, Lütjens, Klemmer, Amara, Steinegger, Zhang, and Zhu]{Reiersen2022}
Gyri Reiersen, David Dao, Björn Lütjens, Konstantin Klemmer, Kenza Amara, Attila Steinegger, Ce~Zhang, and Xiaoxiang Zhu.
\newblock {ReforesTree}: A dataset for estimating tropical forest carbon stock with deep learning and aerial imagery.
\newblock \emph{Proceedings of the 2022 Conference by the Association for the Advancement of Artificial Intelligence}, 2022.

\bibitem[Reiner et~al.(2023)Reiner, Brandt, Tong, Skole, Kariryaa, Ciais, Davies, Hiernaux, Chave, Mugabowindekwe, et~al.]{reiner2023more}
Florian Reiner, Martin Brandt, Xiaoye Tong, David Skole, Ankit Kariryaa, Philippe Ciais, Andrew Davies, Pierre Hiernaux, J{\'e}r{\^o}me Chave, Maurice Mugabowindekwe, et~al.
\newblock More than one quarter of {A}frica’s tree cover is found outside areas previously classified as forest.
\newblock \emph{Nature Communications}, 14\penalty0 (1):\penalty0 2258, 2023.

\bibitem[Rolf(2023)]{rolf2023evaluation}
Esther Rolf.
\newblock Evaluation challenges for geospatial {ML}.
\newblock \emph{Machine Learning for Remote Sening Workshop at the International Conference on Learning Representations}, 2023.

\bibitem[Rolf et~al.(2022)Rolf, Packer, Beutel, and Diaz]{rolf2022striving}
Esther Rolf, Ben Packer, Alex Beutel, and Fernando Diaz.
\newblock Striving for data-model efficiency: Identifying data externalities on group performance.
\newblock \emph{Workshop on Trustworthy and Socially Responsible Machine Learning at Neural Information Processing Symposium}, 2022.

\bibitem[Rolf et~al.(2024)Rolf, Klemmer, Robinson, and Kerner]{Rolf2024}
Esther Rolf, Konstantin Klemmer, Caleb Robinson, and Hannah Kerner.
\newblock Position: Mission critical -- satellite data is a distinct modality in machine learning.
\newblock In \emph{Proceedings of the International Conference on Machine Learning}, 2024.

\bibitem[Roscher et~al.(2024)Roscher, Ru{\ss}wurm, Gevaert, Kampffmeyer, Dos~Santos, Vakalopoulou, H{\"a}nsch, Hansen, Nogueira, Prexl, et~al.]{roscher2024better}
Ribana Roscher, Marc Ru{\ss}wurm, Caroline Gevaert, Michael Kampffmeyer, Jefersson~A Dos~Santos, Maria Vakalopoulou, Ronny H{\"a}nsch, Stine Hansen, Keiller Nogueira, Jonathan Prexl, et~al.
\newblock Better, not just more: Data-centric machine learning for {E}arth observation.
\newblock \emph{IEEE Geoscience and Remote Sensing Magazine}, 2024.

\bibitem[Rußwurm et~al.(2024)Rußwurm, Wang, Kellenberger, Roscher, and Tuia]{Rubwurm2024}
Marc Rußwurm, Sherrie Wang, Benjamin Kellenberger, Ribana Roscher, and Devis Tuia.
\newblock {Meta-learning to address diverse Earth observation problems across resolutions}.
\newblock \emph{Communications Earth \& Environment}, 5\penalty0 (37), 2024.

\bibitem[Saul and Roweis(2003)]{Saul2003}
Lawrence~K. Saul and Sam~T. Roweis.
\newblock Think globally, fit locally: unsupervised learning of low dimensional manifolds.
\newblock \emph{Journal of Machine Learning Research}, 4:\penalty0 119–155, 2003.
\newblock ISSN 1532-4435.
\newblock \doi{10.1162/153244304322972667}.

\bibitem[Stewart et~al.(2022)Stewart, Robinson, Corley, Ortiz, Ferres, and Banerjee]{stewart2022torchgeo}
Adam~J Stewart, Caleb Robinson, Isaac~A Corley, Anthony Ortiz, Juan M~Lavista Ferres, and Arindam Banerjee.
\newblock Torchgeo: deep learning with geospatial data.
\newblock In \emph{Proceedings of the 30th International Conference on Advances in Geographic Information Systems}, pages 1--12, 2022.

\bibitem[Tetemke et~al.(2021)Tetemke, Birhane, Rannestad, and Eid]{tetemke2021species}
Buruh~Abebe Tetemke, Emiru Birhane, Meley~Mekonen Rannestad, and Tron Eid.
\newblock Species diversity and stand structural diversity of woody plants predominantly determine aboveground carbon stock of a dry afromontane forest in northern {E}thiopia.
\newblock \emph{Forest Ecology and Management}, 500:\penalty0 119634, 2021.

\bibitem[Tolan et~al.(2024)Tolan, Yang, Nosarzewski, Couairon, Vo, Brandt, Spore, Majumdar, Haziza, Vamaraju, Moutakanni, Bojanowski, Johns, White, Tiecke, and Couprie]{meta}
Jamie Tolan, Hung-I Yang, Benjamin Nosarzewski, Guillaume Couairon, Huy~V. Vo, John Brandt, Justine Spore, Sayantan Majumdar, Daniel Haziza, Janaki Vamaraju, Theo Moutakanni, Piotr Bojanowski, Tracy Johns, Brian White, Tobias Tiecke, and Camille Couprie.
\newblock Very high resolution canopy height maps from {RGB} imagery using self-supervised vision transformer and convolutional decoder trained on aerial lidar.
\newblock \emph{Remote Sensing of Environment}, 300:\penalty0 113888, 2024.
\newblock ISSN 0034-4257.
\newblock \doi{https://doi.org/10.1016/j.rse.2023.113888}.

\bibitem[Tsao et~al.(2023)Tsao, Nzewi, Jayeoba, Ayogu, and Lobell]{Tsao2023}
Angela Tsao, Ikenna Nzewi, Ayodeji Jayeoba, Uzoma Ayogu, and David~B. Lobell.
\newblock Canopy height mapping for plantations in {N}igeria using {GEDI, Landsat, and Sentinel-2}.
\newblock \emph{Remote Sensing}, 15\penalty0 (21), 2023.
\newblock ISSN 2072-4292.
\newblock \doi{10.3390/rs15215162}.

\bibitem[Vemulapalli et~al.(2023)Vemulapalli, Pouransari, Faghri, Mehta, Farajtabar, Rastegari, and Tuzel]{vemulapalli2023label}
Raviteja Vemulapalli, Hadi Pouransari, Fartash Faghri, Sachin Mehta, Mehrdad Farajtabar, Mohammad Rastegari, and Oncel Tuzel.
\newblock Label-efficient training of small task-specific models by leveraging vision foundation models.
\newblock \emph{arXiv preprint arXiv:2311.18237}, 2023.

\bibitem[Wang et~al.(2023)Wang, Braham, Xiong, Liu, Albrecht, and Zhu]{Wang2023}
Yi~Wang, Nassim Ait~Ali Braham, Zhitong Xiong, Chenying Liu, Conrad~M. Albrecht, and Xiao~Xiang Zhu.
\newblock {SSL4EO-S12}: A large-scale multimodal, multitemporal dataset for self-supervised learning in {E}arth observation [software and data sets].
\newblock \emph{IEEE Geoscience and Remote Sensing Magazine}, 11\penalty0 (3):\penalty0 98--106, 2023.
\newblock \doi{10.1109/MGRS.2023.3281651}.

\bibitem[Weinstein et~al.(2021)Weinstein, Graves, Marconi, Singh, Zare, Stewart, Bohlman, and White]{Weinstein2021}
Ben~G. Weinstein, Sarah~J. Graves, Sergio Marconi, Aditya Singh, Alina Zare, Dylan Stewart, Stephanie~A. Bohlman, and Ethan~P. White.
\newblock A benchmark dataset for canopy crown detection and delineation in co-registered airborne {RGB}, {LiDAR} and hyperspectral imagery from the {National Ecological Observation Network}.
\newblock \emph{PLOS Computational Biology}, 17\penalty0 (7):\penalty0 1--18, 07 2021.
\newblock \doi{10.1371/journal.pcbi.1009180}.

\bibitem[Wilkes et~al.(2015)Wilkes, Jones, Suarez, Mellor, Woodgate, Soto-Berelov, Haywood, and Skidmore]{Wilkes2015}
Phil Wilkes, Simon~D. Jones, Lola Suarez, Andrew Mellor, William Woodgate, Mariela Soto-Berelov, Andrew Haywood, and Andrew~K. Skidmore.
\newblock Mapping forest canopy height across large areas by upscaling {ALS} estimates with freely available satellite data.
\newblock \emph{Remote Sensing}, 7\penalty0 (9):\penalty0 12563--12587, 2015.
\newblock ISSN 2072-4292.
\newblock \doi{10.3390/rs70912563}.

\bibitem[Yeh et~al.(2020)Yeh, Perez, Driscoll, Azzari, Tang, Lobell, Ermon, and Burke]{Yeh2020}
Christopher Yeh, Anthony Perez, Anne Driscoll, George Azzari, Zhongyi Tang, David Lobell, Stefano Ermon, and Marshall Burke.
\newblock \href{https://www.nature.com/articles/s41467-020-16185-w}{{Using publicly available satellite imagery and deep learning to understand economic well-being in Africa}}.
\newblock \emph{Nature Communications}, 11\penalty0 (2583), 2020.

\bibitem[Zhou et~al.(2003)Zhou, Bousquet, Lal, Weston, and Sch\"{o}lkopf]{Zhou2003}
Dengyong Zhou, Olivier Bousquet, Thomas Lal, Jason Weston, and Bernhard Sch\"{o}lkopf.
\newblock Learning with local and global consistency.
\newblock In \emph{Advances in Neural Information Processing Systems}, volume~16, 2003.

\end{thebibliography}

\newpage

\appendix
\section{Additional supporting results}
\label{app:extra_results}
This appendix details supporting figures and tables referenced from the main text. 

\begin{figure}[h]
    \centering
    \includegraphics[width=1\textwidth]{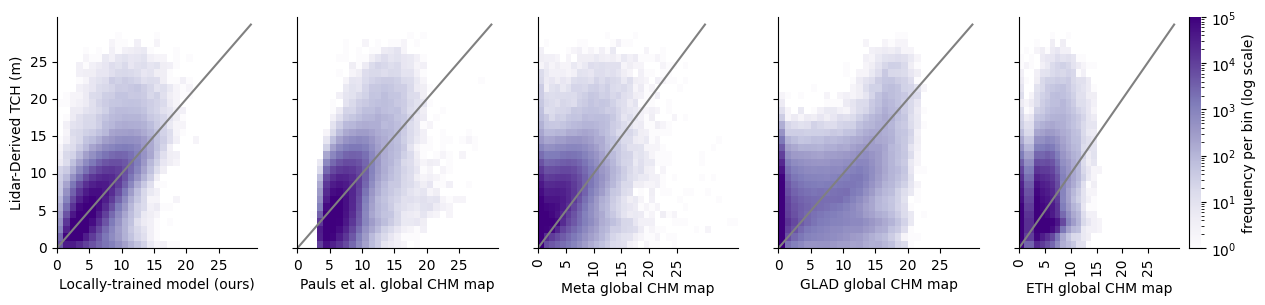}
    \caption{Distribution of predictions and errors from each model (horizontal axis) vs. the LiDAR-derived ground-referenced measures of TCH, binned to the nearest 1m.}
    \label{fig:prediction_distributions}
\end{figure}

\begin{figure}[h]
    \centering
    \includegraphics[width=\textwidth]{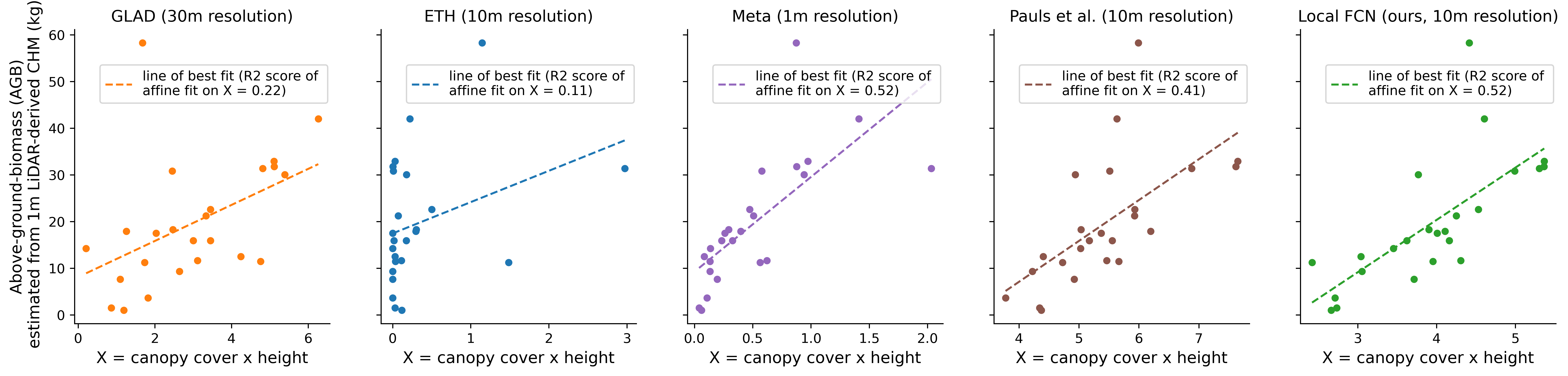}
    \caption{Aboveground biomass (AGB) per site, as estimated using the 1m LiDAR-derived TCH labels and the linear formula $X = CC \times H$ from \cite{Colgan2013} (where $CC=$ canopy cover and $H$=height per pixel, where reference values are computed on label data with 1m resolution). Specifically, equation F “PA harvest” from \cite{Colgan2013} is calculated for each site using our high-fidelity LiDAR-derived local data at 1m resolution, and plotted on the vertical axis. On the horizontal axis of each panel, we plot the same ``$X = CC \times H$'', this time estimated from each of the four different data products and our own predictive model. For the GLAD map, the ETH map, the map from \cite{pauls2024estimating}, and the map from our local model, we use 10m resolution. The Meta map has 1m resolution. Dashed lines show the line of best fit describing the least-squares linear fit of that variable to the AGB values at each site calculated from our labels. The goodness-of-fit of each line is measured by the $R^2$ score of a linear affine projection of the form $X\beta_1 + \beta_0$.}
    \label{fig:agb_comparison}
\end{figure}

\label{appendix:xception_layers}
\begin{table}[ht]
\centering
\resizebox{\textwidth}{!}{
\begin{tabular}{llcccccc}\toprule
& & \multicolumn{2}{c}{RMSE} &\multicolumn{2}{c}{MAE} & \multicolumn{2}{c}{R2} \\
\cmidrule(lr){3-4}\cmidrule(lr){5-6}\cmidrule(lr){7-8}
model description & layers tuned  & average & std dev. & average & std dev. & average & std dev. \\\midrule
 \multirow{3}{*}{XcpetionS2 (fine-tuned)} & last 1 & 2.09 & 0.02 & 1.58 & 0.02 & 0.30 & 0.01 \\
 & last 2 & 1.97 & 0.03 & 1.47 & 0.02 & 0.38 & 0.02 \\
& last 3 & 1.98 & 0.03 & 1.48 & 0.02 & 0.37 & 0.02 \\
\bottomrule
\multirow{3}{*}{XcpetionS2 (random init.)  no latlon} & last 1 & 2.21 & 0.02 & 1.72 & 0.02 & 0.22 & 0.02 \\
& last 2 & 1.97 & 0.03 & 1.49 & 0.03 & 0.38 & 0.02\\
& last 3 & 1.94 & 0.03 & 1.47 & 0.02 & 0.40 & 0.02\\
\bottomrule
 \multirow{3}{*}{XcpetionS2 (random init.)}& last 1 & 2.31 & 0.05 & 1.82 & 0.04 & 0.15 & 0.03\\
& last 2 &2.08 & 0.05 & 1.58 & 0.04 & 0.31 & 0.03\\
& last 3 &2.05 & 0.04 & 1.55 & 0.03 & 0.33 & 0.03\\
\bottomrule
\end{tabular}}
\caption{Performance after fine-tuning the last 1, 2, or 3 layer(s) of an XceptionS2 model from \cite{eth} with our local data. There are small gains to tuning 2 vs 3 layers. 
This extends the results from  \Cref{table:fine_tuning}, which shows performance after tuning the last 3 layers in a transfer-learning approach. }
\label{table:xception_fine_tuning_layers}
\end{table}

\ifx\jmlrsubmissionversion\truecondition
{}
\else{
\def\subfigheight{2in}
\def\croptrim{1.1}
\begin{figure*}[t!]
    \centering
    \begin{subfigure}[t]{1.0\textwidth}
        \centering
        \includegraphics[trim={0 \croptrim cm 0 0}, clip, height=\subfigheight]{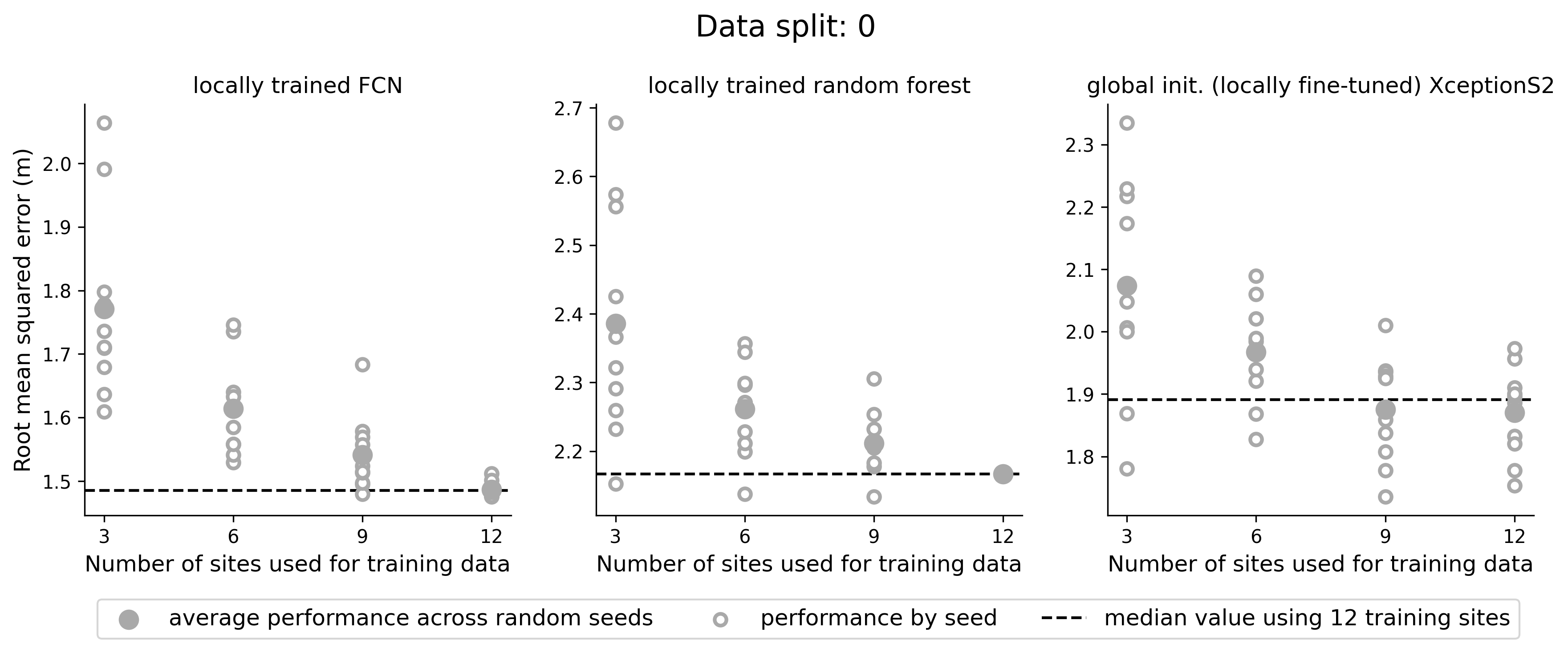}
    \end{subfigure} \\
    \begin{subfigure}[t]{1.0\textwidth}
        \centering
        \includegraphics[trim={0 \croptrim cm 0 0}, clip, height=\subfigheight]{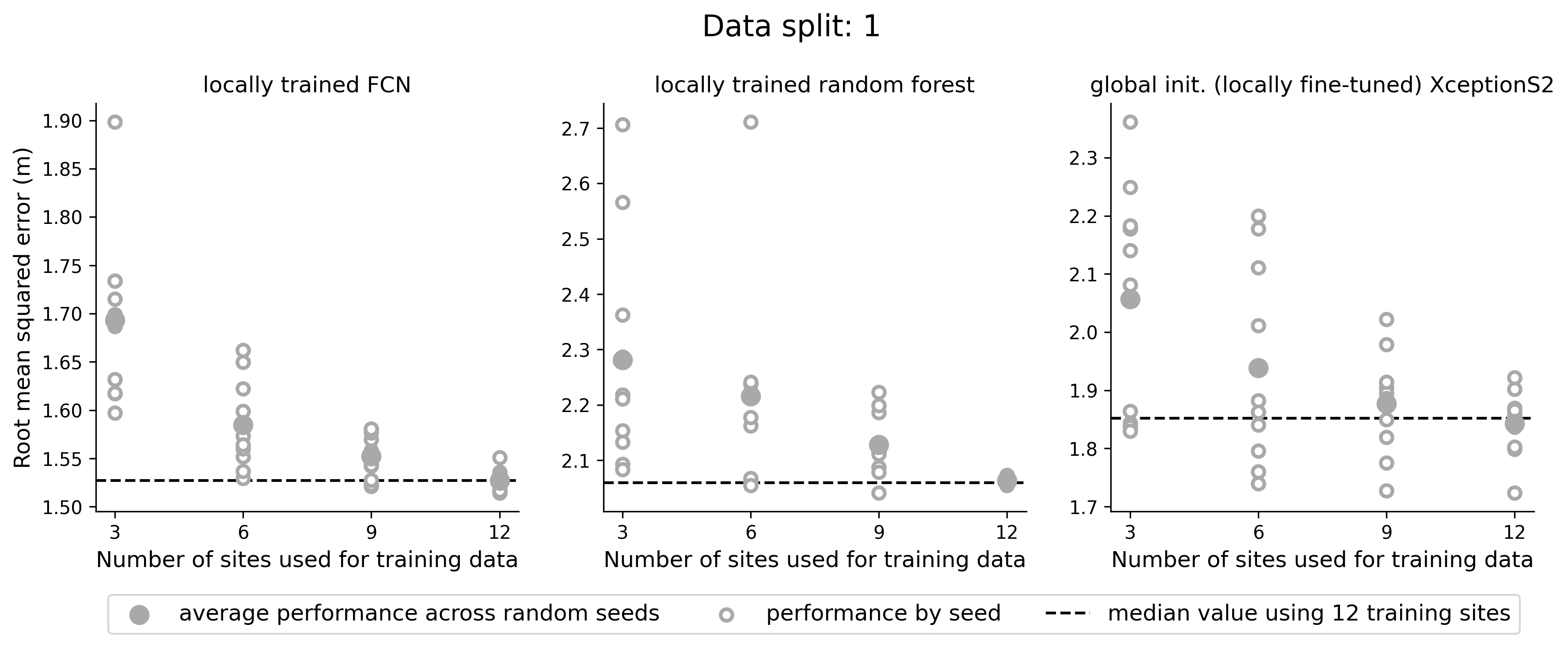}
    \end{subfigure} \\
    \begin{subfigure}[t]{1.0\textwidth}
        \centering
        \includegraphics[trim={0 \croptrim cm 0 0}, clip, height=\subfigheight]{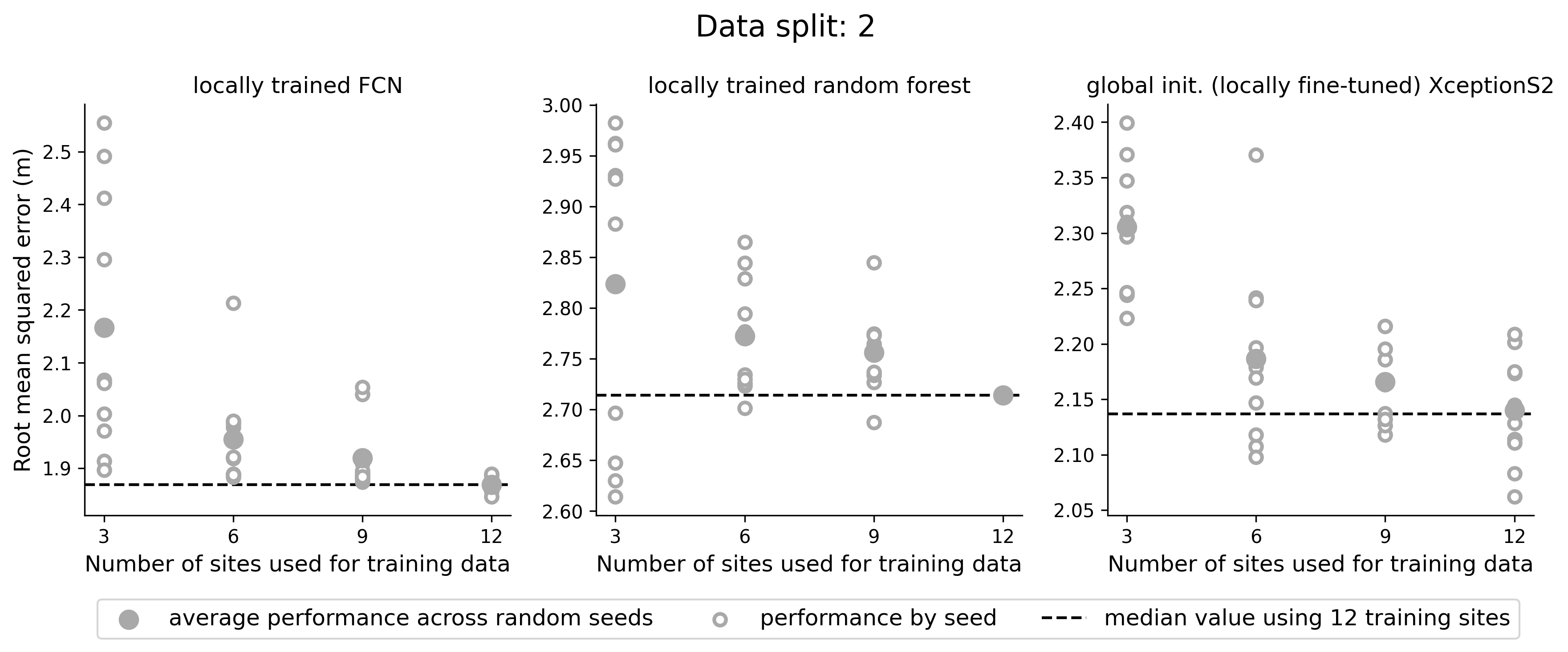}
    \end{subfigure} \\
    \begin{subfigure}[t]{1.0\textwidth}
        \centering
        \includegraphics[height=\subfigheight]{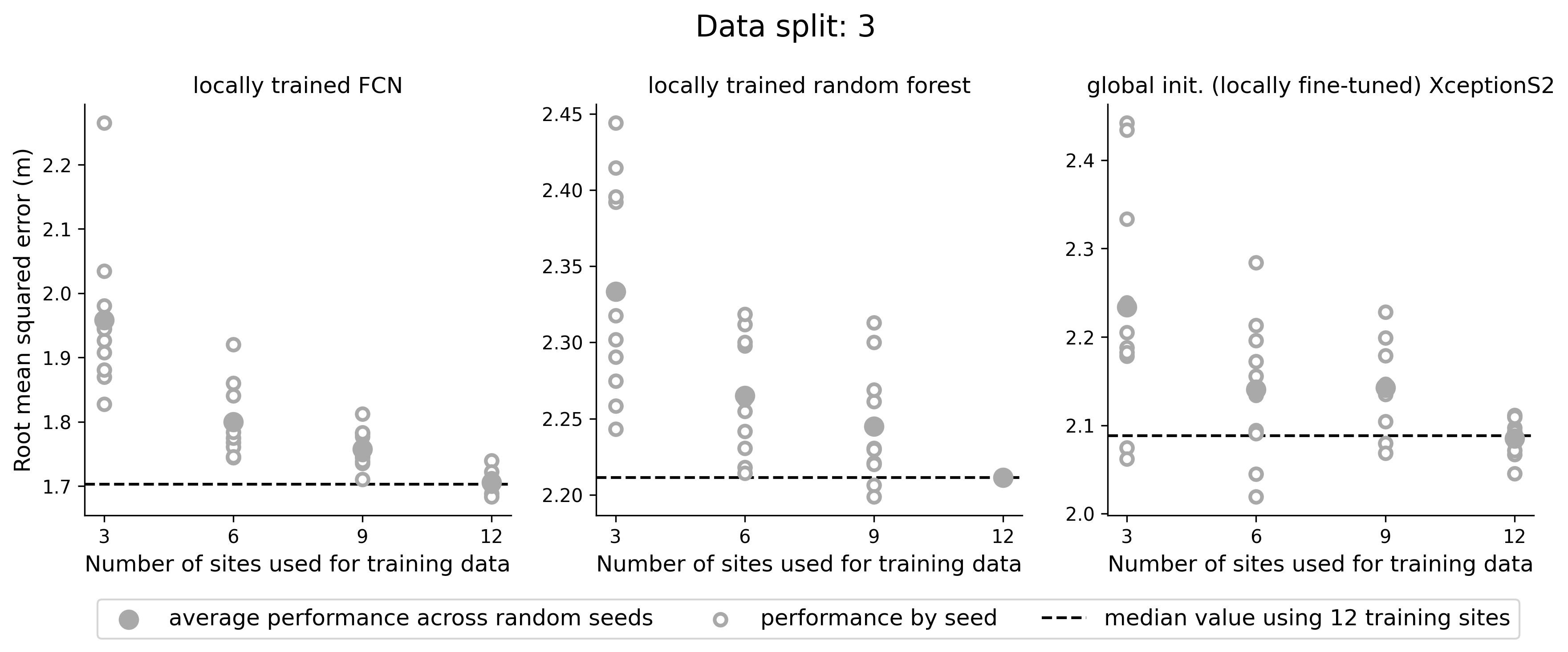}
    \end{subfigure}
    \caption{Results extending \Cref{fig:data_model_interactions} for all four data splits.
    \label{fig:data_model_interactions_all_splits}}
\end{figure*}
}
\fi

\begin{table}
\centering
\resizebox{\textwidth}{!}{
\begin{tabular}{llcc}\toprule
model description & layers tuned & trainable parameters & total parameters
\\\midrule
XceptionS2 & last 1 &  257 & 1,731,587 \\
XceptionS2 & last 2 &  203,265 & 1,731,587  \\
XceptionS2 & last 3 &  406,273 & 1,731,587  \\
U-Net (ResNet-18 backbone) & decoder only &  3,151,697 & 14,356,433 \\
U-Net (ResNet-18 backbone) & all &  14,356,433 & 14,356,433 \\
5 layer FCN (``ours'') & all &  604,417 & 604,417 \\
\bottomrule
\end{tabular}}
\caption{Number of total parameters and trainable parameters for the neural network models that we compare in \Cref{table:fine_tuning,table:xception_fine_tuning_layers}, assuming 12 input channels.}
\label{table:num_params}
\end{table}

\section{Evaluation procedure for stratifying by feature}
\label{appendix:stratifying-by-feature}
\subsection{Ground-truth Height}
To stratify performance by the ground-truth canopy height, we calculate the residuals for the pixels whose label falls within the specified range.
\subsection{Distance to River}
To stratify performance by distance to river, we start with a data product provided to us by researchers at the Karingani Game Reserve spanning the study area in vector format that has rivers encoded as lines. Note that these ``rivers'' correspond to paths through which water could flow but are not necessarily active rivers during all parts of the year. We rasterize this vector data and then for each site, we calculate the distance between each pixel in the site and every river pixel. We then take the minimum of those as the ``distance to river'' for that pixel. During evaluation we then calculate the residuals for the pixels whose minimum distance to any river falls within the specified range.
\subsection{Geology}
To stratify performance by geology type, we start with a data product spanning the study area in vector format that has geology type encoded as polygons, provided to us by researchers at the Karingani Game Reserve. We rasterize this vector data and then for each site, we record the geology type for each pixel in the site. During evaluation we then calculate the residuals for the pixels whose geology type matches the specified category.
\end{document}